\documentclass[preprint,12pt]{elsarticle}

\usepackage{amssymb}
\usepackage{amsmath}

\journal{Applied Soft Computing}

\usepackage{graphicx}%
\usepackage{multirow}%
\usepackage{amsmath,amssymb,amsfonts}%
\usepackage{amsthm}%
\usepackage{mathrsfs}%
\usepackage[title]{appendix}%
\usepackage{xcolor}%
\usepackage{textcomp}%
\usepackage{manyfoot}%
\usepackage{booktabs}%
\usepackage{algorithm}%
\usepackage{algorithmicx}%
\usepackage{algpseudocode}%
\usepackage{listings}%
\usepackage{hyperref}

\usepackage{cleveref}
\usepackage{tabularx,ragged2e,caption}

\newcounter{stepnum} 

\Crefformat{figure}{#2figure~#1#3}
\Crefformat{figure}{#2Figure~#1#3}
\Crefformat{algorit}{#2Algorithm~#1#3}
\Crefformat{algorit}{#2algorithm~#1#3}
\Crefformat{algocf}{#2Algorithm~#1#3}
\Crefformat{algocf}{#2algorithm~#1#3}
\Crefformat{table}{#2table~#1#3}
\Crefformat{table}{#2Table~#1#3}
\Crefformat{equation}{#2equation~#1#3}
\Crefformat{equation}{#2Equation~#1#3}
\Crefformat{section}{#2section~#1#3}
\Crefformat{section}{#2Section~#1#3}
\Crefformat{chapter}{#2chapter~#1#3}
\Crefformat{chapter}{#2Chapter~#1#3}

\definecolor{RED}{RGB}{255,0,0}

\usepackage[latin1,utf8]{inputenc}
\usepackage{rotating}
\usepackage{nomencl}
\usepackage{subcaption}

\begin{document}


\begin{frontmatter}



\title{Balancing Performance and Reject Inclusion: A Novel Confident Inlier Extrapolation Framework for Credit Scoring}


\author[1]{Athyrson M. Ribeiro\corref{cor1}}
\ead{athyrson@ic.unicamp.br}
\author[1]{Marcos Medeiros Raimundo}
\ead{mraimundo@ic.unicamp.br}
\affiliation[1]{
    organization = {University of Campinas},
    adressline = {Av. Albert Eistein, 1251},
    postcode = {13083-852},
    city = {Campinas},
    country = {Brazil}
}

\cortext[cor1]{Corresponding author}

\begin{abstract}
Reject Inference (RI) methods aim to address sample bias by inferring missing repayment data for rejected credit applicants. Traditional approaches often assume that the behavior of rejected clients can be extrapolated from accepted clients, despite potential distributional differences between the two populations. To mitigate this blind extrapolation, we propose a novel Confident Inlier Extrapolation framework (CI-EX). CI-EX iteratively identifies the distribution of rejected client samples using an outlier detection model and assigns labels to rejected individuals closest to the distribution of the accepted population based on probabilities derived from a supervised classification model. The effectiveness of our proposed framework is validated through experiments on two large real-world credit datasets. Performance is evaluated using the Area Under the Curve (AUC) as well as RI-specific metrics such as Kickout and a novel metric introduced in this work, denoted as Area under the Kickout. Our findings reveal that RI methods, including the proposed framework, generally involve a trade-off between AUC and RI-specific metrics. However, the proposed CI-EX framework consistently outperforms existing RI models from the credit literature in terms of RI-specific metrics while maintaining competitive performance in AUC across most experiments.
\end{abstract}




\end{frontmatter}




\section{Introduction}\label{sec1}
Credit Scoring affects the vast majority of people worldwide. Whenever a client requires a loan (or a credit card), the bank calculates their credit score to evaluate their ability to pay their debts \cite{mester1997s}. The fear of granting loans to many \textit{default} applicants leads companies to use harsh credit scoring policies. Such lending standards could exclude many good debtors and exacerbate harm to under-represented communities. A credit model trained with only a diminutive and under-representative subset of society is not ideal for classifying the whole population reasonably and precisely. This phenomenon is known as sample bias, and it occurs when a credit model is trained with only accepted clients \cite{guo2023transductive, nikita_kozodoi_shallow_2019, kang2021graph}. Due to the harsh policies the companies apply, most applicants are denied a loan, meaning the rejected applicants constitute the majority of data in credit datasets\cite{kang2021graph, shen2020three}. Some approaches assimilate information from rejected and accepted clients to improve credit scoring systems.
This group of techniques is called Reject Inference (RI). The incorporation of RI techniques grants substantial advantages: (1) A considerable decrease of sample bias --- coming from more robust models of credit scoring trained with information of a more significant population; (2) Minimization of data waste; (3) Better evaluation of marginalized communities.

RI literature has advanced considerably in the last few decades, and many papers have been published highlighting the importance of RI application in the credit scoring process. From simple assumptions, considering all rejected as bad cases (potential defaults) to an entire network using rejected clients' information to infer credit scoring \cite{siddiqi2017intelligent, liu2022rmt}. However, some strong assumptions in RI literature can not be ignored. The first one is that the behavior of the rejected population can be extrapolated based on the accepted population. This is often not the case, as there are many differences in the distribution of accepted and rejected clients. The second assumption is that a slight gain in accuracy is the objective of RI applications. When the entire pipeline, from training to testing, is based solely on the accepted population, credit scoring models can already have high predictive accuracy. However, we believe ignoring the existence of sample bias is not a good way to tackle credit scoring, as many people historically outside the distribution of the accepted population can be harmed. 

Many recent papers in RI literature propose frameworks combining several RI and machine learning techniques to label and filter out samples. This combination seems to lead to models with high classification power \cite{shih_framework_2022, shen2020three, liao_combating_2022}. However, most RI literature is based on at least one of the previous assumptions. This research proposes a novel framework with several verification steps to ensure confidence in the RI process utilized. Confident-Inline Extrapolation for Rejection Inference (CI-EX) uses outlier detection and classification probabilities to label and filter the most confident samples. The framework is built on an iterative procedure, where each iteration implies a new model that is more aware of the RI population distribution than its predecessor. This is made to avoid the extrapolation bias. We tackle the assumption about accuracy in RI by using metrics that consider the RI population. We argue that these metrics are more suited to evaluate the actual performance of RI techniques. We evaluate our method using the Reject Inference metric Area under Kickout (AUK), based on the kickout metric introduced by \cite{kozodoi2020shallow}. This metric was explicitly designed for RI scenarios due to its stronger correlation with correctly assessing the unbiased population. Our proposed framework consistently outperforms other Reject Inference techniques in the literature on this RI-specific metric. More specifically, our contributions are:

\begin{itemize}
    \item We introduce a novel semi-supervised framework, Confident-Inline Extrapolation for Rejection Inference (CI-EX), which effectively combines outlier detection and confidence-based selection to enhance the accuracy of class predictions for rejected samples.
    \item We propose the Area Under the Kickout (AUK) metric, a new and unbiased performance measure specifically designed for evaluating Reject Inference models, addressing a gap in current model assessment practices.
    \item We provide a comprehensive review, evaluation, and implementation of classical Reject Inference models from the literature. We apply metrics that account for accepted and rejected clients, offering a more holistic view of model performance in real-world RI scenarios.
\end{itemize}

\section{Literature Review}

\subsection{Credit Scoring}

Credit scoring is critical to many processes in granting loans, leasing properties, and other commodities. The decision to approve or to deny a loan to a borrower hinges on their ability to convincingly assure the lender of their trustworthiness \cite{anderson2022credit}. However, if this decision is made without a protocol or transparency, many problems can arise. The most obvious problem is the financial loss caused by lending funds to borrowers who will not repay them. They are traditionally called bad payers in credit scoring literature (the borrowers who pay back on time are called good payers). Therefore, implementing an automatic, or at least semi-automatic, trustworthiness system is crucial. This system is known as credit scoring \cite{kang2021graph}. For simplicity, without loss of generality, from now on, we will limit our discussion to the process of credit scoring that involves a company lending funds to an individual.

The credit scoring process generally involves obtaining information about an individual and comparing this information to other individuals, from which we have payment behavior data. In machine learning, this information about an individual is called features, and the classification of whether the individual is a good or bad payer is called class or target. The assumption is that the payment behavior of an individual can be estimated based on their features. The features that may assist in this estimation are often related to the client's economic situation, the loan itself, or the individual's historical credit data (which, in many cases, is unavailable to the company). With the use of these features and respective targets, classification models can be fitted by the company to assist in the process of selecting trustworthy clients to grant loans. 

\subsection{Reject Inference}

\begin{figure}[htbp]
    \centering
    \includegraphics[width=0.8\linewidth]{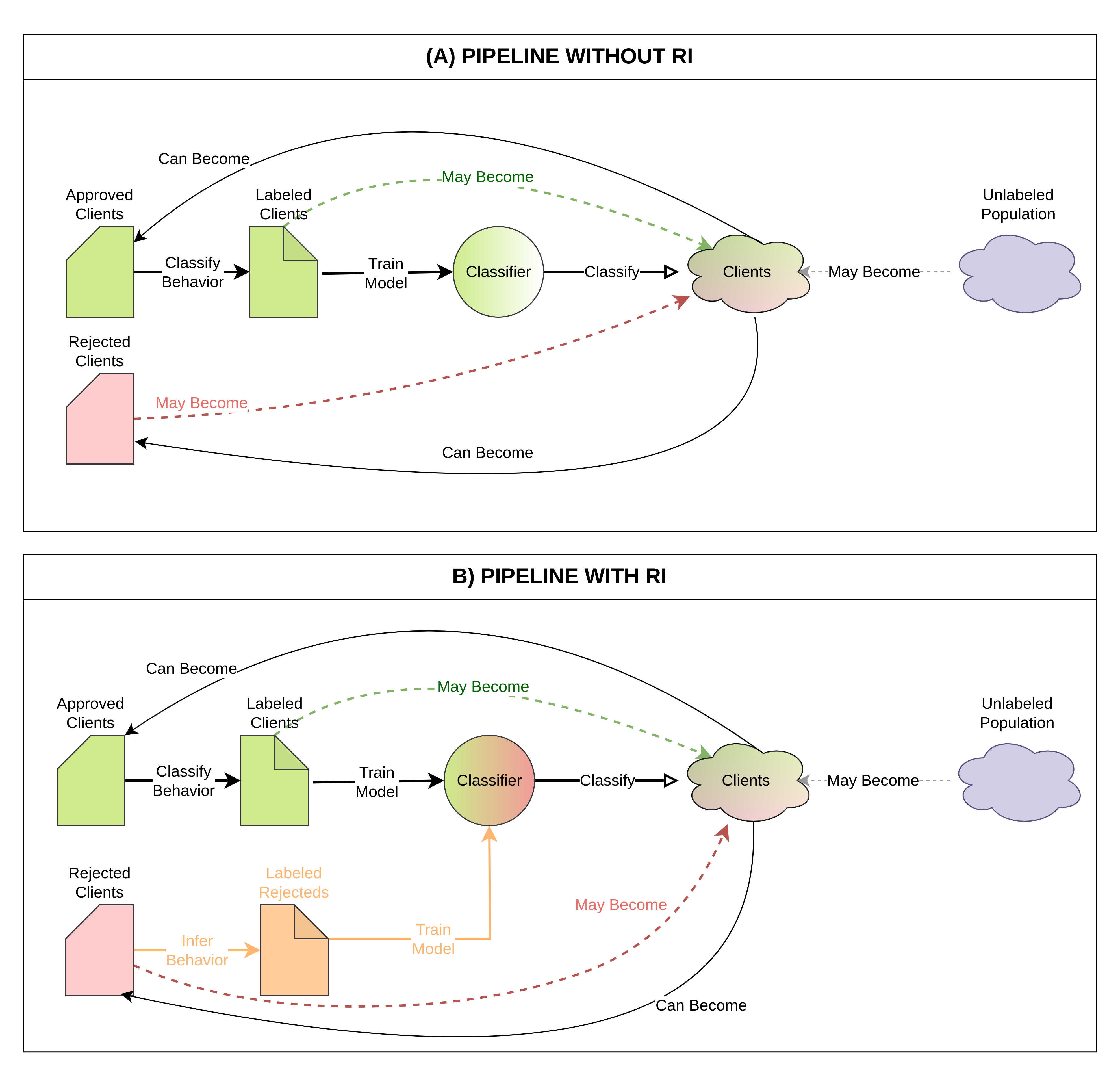}
    \caption{(A) Pipeline that discards rejected clients data, versus (B) pipeline that applies Reject Inference.}
    \label{fig:TwoPipes}
\end{figure}

When building a classifier to automate the decision of who should be worthy of receiving a loan, an essential requirement is that such a classifier is good at generalization. In realistic terms, such a model should perform well even with data that differs, to some extent, from the data it was trained upon. When training Machine Learning models, we separate the data into training, validation, and testing. The model's generalization directly relates to how much the data it was fitted reflects the real world in which it will be applied. Therefore, when only data from accepted clients is used in training the credit pipeline, as illustrated on \Cref{fig:TwoPipes} a), a clear sample bias is identified. \Cref{fig:TwoPipes} a) illustrates a credit pipeline from a company that builds its classification model based only on approved clients from previous iterations. However, not only approved clients but also the population rejected by earlier iterations and clients coming from unseen distributions may ask for a loan from this company. Therefore, we have a model based on a sample that does not accurately reflect the entire population, resulting in what is known as sample bias. At each iteration of this pipeline, the sample bias will only grow, leading the company to use models of classification that are less applicable to the entire population each time \cite{siddiqi2017intelligent}.

The biggest obstacle in avoiding sampling bias in credit scoring is the lack of labels for the rejected clients. Approved clients, as illustrated in \Cref{fig:TwoPipes}, can have their behavior observed. For example, depending on whether they repay the loan within the stipulated time, they can be almost accurately classified as good or bad payers. The same cannot be applied to rejected clients. The company had their data when they asked for the loan but can not retrieve accurate labels based on their repayment behavior. One solution would be to approve all clients and classify them according to their behavior \cite{anderson2022credit}. However, this would be too costly for most loan companies. Luckily, there are more applicable solutions from both the literature and the business. These solutions are known as Reject Inference (RI), and most of them can be described as making the classification model aware of the rejected client population. \Cref{fig:TwoPipes} b) illustrates RI in a credit pipeline. As shown in the \Cref{fig:TwoPipes} b), the behavior of the rejected clients is inferred through some technique, and a label is given to them based on that. The data from these clients is then concatenated to the training set to build the new classifier. The model resulting from this process is a model that has more knowledge of the whole population than the model built with only accepted clients. 

However, Reject Inference is not without flaws and caveats. First, it should be mentioned that there are other approaches to the technique other than using it to inflate the training set, some of which will be described in the following subsections. Second, RI and statistical processes are built on a series of assumptions, such as the type of missing data problem, the viability of inferring the missing features and labels of rejected clients, and the evaluation process capable of measuring the actual performance of the credit pipelines.  

Reject Inference (RI) techniques vary significantly in incorporating the rejected data into the credit model pipeline. Even RI techniques in the same family can have very different approaches, as for augmentation techniques \cite{siddiqi2017intelligent}. Because of such diversity, there is no consensus on the best technique for all scenarios. Each technique also has flaws and restraints \cite{crook_does_2004}. Despite their limitations, the application of an RI technique should bring a credit scoring model that is more robust, less biased toward the whole population, and less wasteful of data. 

Many authors \cite{el2022semihmms, shen2020three, liu2022rmt, anderson2022credit} mention the three types of missingness of data proposed by \cite{little2019statistical} when introducing the lack of rejected data in most credit scoring systems \cite{el2022semihmms}:
\begin{itemize}
    \item Data can be missing due to completely random reasons (MCAR) when there is no relation between the missingness of the data and any other variable related to the system or sample;
    \item Data can be missing at random (MAR) when there is a relation between the missingness of the variable of interest and some other variable in the dataset that is not the variable of interest;
    \item Data can also be missing not at random (MNAR), when the missingness is related to the missing data itself and may be caused by some unobserved variables. 
\end{itemize}

According to \cite{liu2022rmt}, MNAR can play a significant role in RI due to the subjective reasons that can influence the approval of a loan in not fully automated credit scoring systems. \cite{anderson2022credit} also affirms that most cases of missing data in credit scoring systems can be attributed to MNAR due to the outside factors that can not be represented in a credit model but influence the decision of which applicants will be rejected. 

The RI technique can be applied in different stages of the model pipeline. Maybe the most intuitive approach would be to infer the labels of the rejected clients to expand the training set with their data eventually, like extrapolation \cite{siddiqi2017intelligent}, parceling \cite{siddiqi2017intelligent}, and label spreading \cite{zhou2003learning}: the Data Inflating Methods. Some techniques, however, only apply the rejected data in the form of adjusting the weights of the credit model, which is the case for most types of augmentation \cite{siddiqi2017intelligent, anderson2022credit}: the Weight Adjusting Methods. Some authors go a step further and propose new machine learning models built to consider the existence of rejected data \cite{liu2022rmt}: the Model Approach Methods. 

\cite{siddiqi2017intelligent} explains that the usefulness of RI techniques is highly linked to our confidence in our previous system for the Approval/Rejection of loans. RI may not be indicated if the confidence is too low, close to decided randomly, or too high, with a high approval rate. Although it is not a recommended strategy, if the confidence is too high, one straightforward RI technique that can be applied is to assume all rejects as bad payers \cite{siddiqi2017intelligent, anderson2022credit}. There are, however, many reasons for the application of RI techniques. The most common reason is to avoid sample bias by using a subset not truly representative of the whole population \cite{siddiqi2017intelligent, kang2021graph, song2022towards, shen2020three}. Another strong reason for applying RI techniques is to fix past decisions made in credit scorecard development. For example, RI can help make marginalized individuals more considered and less prejudiced in the credit process \cite{siddiqi2017intelligent}. For financial institutions, RI can inform a more accurate default rate of the population, avoiding monetary losses \cite{liaoDataAugmentationMethods2021, siddiqi2017intelligent}.  The following subsections describe the three RI techniques mentioned in this section.

\subsubsection{Weight Adjusting Methods}

\label{Weighttechniques}

Augmentation, also known as Reweighing, is a technique where the weights of the accepted data are adjusted to consider the probabilities of rejection \cite{siddiqi2017intelligent, anderson2022credit}. An approval/rejection (AR) model is fitted with an accept and reject status and is used as the class. The model is then applied to the accepted data, and each sample's probability is retrieved. In Upward Augmentation (A-UW), the new weight is calculated by \Cref{eq:UA}, while in Downward Augmentation (A-DW), the new weight is calculated by \Cref{eq:DA}. Where $\hat{w}$ is a new weight, $w$ is the previous weight (we can assume one as its value), and $p(A)$ is the probability of being accepted given by the AR model \cite{anderson2022credit}.

\begin{equation}
  \hat{w} = \frac{w}{p(A)}
  \label{eq:UA}
\end{equation}
\begin{equation}
\hat{w} = w \cdot (1 - p(A)) 
\label{eq:DA}
\end{equation}

Another way to use Augmentation is to sort the accepted and rejected samples by the $p(A)$, then separate these samples into $n$ splits according to the $p(A)$. For each split, the proportion of accepts between accepts and rejects contained in that split is calculated ($AF = \frac{nA}{nA + nR}$). Then, the augmentation factor for that split will be $\frac{1}{AF}$. The AF will then be used as the new weight for all accepted samples in that split. This technique is called Augmentation with Soft Cut-Off (A-SC) \cite{siddiqi2017intelligent, ehrhardt2021reject}. One more well-known Augmentation method is Fuzzy-Augmentation (A-FU), also known as Fuzzy-Parceling \cite{anderson2022credit}. A key differentiator of this technique is that it is both a Data Inflating Method and a Weight Adjusting Method. In this technique, an AR model is also fitted; however, the rejected data is concatenated to the new dataset twice. First, it is appended receiving $0$ as a label and $p(A)$ as weight, and then it is again appended but with $1$ as a label and $p(R)$ (probability of rejection) as weight. The accepted samples receive $1$ as weight.

\subsubsection{Data Inflating Methods}
\label{Inflatingtechniques}

The use of information about the labeled (accepted) data to infer the labels of the non-labeled (rejected) data is known as Extrapolation \cite{anderson2022credit}. Simple extrapolation techniques use a classifier fitted to the accepted data to infer the labels of the rejected data. Suppose we assume our classifier is good enough to do this inference process. In that case, we can use the inferred labels for the rejected samples as actual labels and concatenate the rejected samples in the new training set. However, it may not be wise to append all the rejected samples simultaneously to the new training set. If we are interested in balancing the number of bad payers in the training set, we could, for example, add only the samples inferred as bad payers from the rejected group; we will call this alternative "Bad Extrapolation" (BE). Another choice would be to consider our confidence in the predictions of our extrapolation model, and to add only the samples farthest from the classification threshold; we will call this alternative "Confident Extrapolation" (E-C). 

Instead of using a fitted classifier to infer the labels of the rejected data, we can infer the labels of the rejected data alongside the training of a label-spreading classifier. Proposed by \cite{zhou2003learning}, this technique relies upon the assumption that nearby samples in a dataset are inclined to have the same labels. After the label spreading classifier is fitted, we can retrieve the labels attributed to the rejected samples by the model. Then, we can expand the training set by concatenating the rejected samples labeled by the label spreading classifier to the training data. We will abbreviate this technique as LSP.

Parcelling (PAR) \cite{siddiqi2017intelligent} is a technique similar to ASC. However, instead of changing the accepted weights, we use the splits to label the rejected samples in this technique. First, a classifier is fitted with accepted data. This classifier is then used to calculate the probability of default on both accepts and rejects. These samples are then sorted based on their probability of default and split based on score intervals. The number $n$ of score intervals is an arbitrary parameter. For each split, we calculate the ratio of actual bad payers (${\beta}$) between all accepted included in that split. But, since we are interested in labeling the rejects, we multiply the bad rate by a prejudice factor $\rho$. With the updated bad rate ($\hat{\beta}$), we can calculate the new expected good rate: $\hat{\kappa} =  1 - \hat{\beta}$. The rejected samples in the split are then randomly assigned a label in proportion to the updated good and bad rates for that split. Once this process is concluded for all splits, the rejected samples can be concatenated to the new training set. 

\subsubsection{Model Approach Methods}
A more recent approach to RI is the creation of machine learning models that are specifically designed to work with both accepts and rejects. In their work, \cite{liu2022rmt} propose a Reject Aware Multi-Task Network (RMT-Net) that takes into consideration the high correlation between the tasks of classification between approval/rejection and default/non-default clients to improve its learning capabilities. Another RI network, proposed by \cite{guo2023transductive}, Transductive Semi-Supervised Metric Network (TSSMN) consists of the union of two networks, the first one is responsible for mapping the samples into a metric space. The second one uses transductive label propagation to label the samples according to the proximity given by the first network. 

\subsection{Outlier Detection}
Outlier Detection is a relevant concept in machine learning. An outlier is a sample that differs too much from the samples of a distribution, which implies it does not belong to that distribution. Subsequently, an inlier is seen as a sample that belongs to that distribution on which the outlier detection (OD) algorithm was trained \cite{xia_novel_2019, ali2023novel}. Generally, removing outliers from the training dataset is expected to translate to a model's higher performance. Therefore, the OD models, such as Isolation Forest \cite{liu2008isolation}, are usually employed to identify outlier samples that should be removed from the dataset. However, some authors have found OD as a tool for more ambitious tasks \cite{xia_novel_2019, nikita_kozodoi_shallow_2019,coenen_probability_2020}. 

Since data for rejected clients does not contain ground truth labels, OD algorithms are well-suited for RI techniques because most are based on unsupervised learning, which does not require labels for training \cite{xia_novel_2019}. In their work, \cite{xia_novel_2019} proposed using OD as a Data Inflating Method for RI. They use Isolation Forest to label samples in the rejected dataset. Outliers in the rejected dataset are seen as samples that should not have been rejected and are reclassified as suitable applicants. The inliers are seen as correctly rejected samples and should be classified as bad applicants. The authors claim to be the first to employ OD as a RI technique, and their work inspired others.

Another combination of OD and RI techniques is found in the works of \cite{nikita_kozodoi_shallow_2019,  coenen_probability_2020} and more recently in \cite{shih_framework_2022}. \cite{nikita_kozodoi_shallow_2019} used OD to iteratively identify inadequate samples from the rejected dataset based on the distribution of the accepted population, ignoring those samples too close and too far from the accepted population. Where \cite{coenen_probability_2020} approach was to use OD to reclassify samples from both the accepted and rejected population in the pre-processing stage. Accepted samples marked as outliers were removed from the accepted dataset, and outliers in the rejected population were incorporated into the training set as suitable applicants. \cite{shih_framework_2022} followed an approach more similar to \cite{xia_novel_2019}, however. In their work, OD was applied to identify potential good cases between the rejected population and remove potential bad cases from the accepted population, effectively using OD for relabeling samples.

\subsection{Related Work}
In their work, \cite{xia_novel_2019} proposed one of the first applications of outlier detection in RI. Unlike most works at the time, they applied outlier detection after the pre-processing phase of the pipeline to label rejected samples. However, although the authors criticized previous literature assumptions on the direct extrapolation of behaviors from the accepted to the rejected population, they also applied outlier detection with a similar principle. Labeling all outliers of the rejected group as good payers, they assumed the entire rejected population could be directly divided between good and bad payers. However, according to \cite{coenen_probability_2020}, not all samples from the rejected group can be reliably labeled based on, i.e., there will be some cases where there will not be enough information to infer the label of a sample based on its features.  Besides, the pre-defined contamination threshold will influence the number of individuals selected as outliers in the reject set. If this is the only criterion utilized, the number of inferred good payers between the rejected population can be vastly exaggerated. Despite that, their work achieved great results, surpassing the models trained with only accepted samples, and influenced others in the RI literature \cite{coenen_probability_2020, shih_framework_2022}.

More recently, \cite{shih_framework_2022} proposed a similar application of outlier detection for RI. Adding to the method proposed by \cite{xia_novel_2019}, the authors ruled that outliers among rejected samples would be classified as good payers. In contrast, outliers in accepted samples should be excluded from the training set (as rejects). Another contribution from the authors was using K-nearest neighbor to fill in missing features in the rejected dataset, addressing the significant discrepancy between the number of features in the accepted and rejected datasets. The authors achieved great results from this combination of techniques with their proposed framework for the Lending Club dataset. However, the authors only measured the performance of their techniques solely on accepted samples and utilized features that could only be obtained after the loan approval phase. 

In their work, \cite{liaoDataAugmentationMethods2021} applied a self-training method for RI where rejected samples would be iteratively added to the training set and labeled based on their prediction confidence. The authors claim that labeling data with low certainty has a low chance of improving classification performance. With this, the authors were able to augment the training dataset by $ 126\%$ and achieve better results in the majority of experiments than the model trained with approved-only samples and other RI methods studied. However, the authors demonstrated their results only with one private and relatively small dataset. 

Outlier detection, as an unsupervised method, has great potential in reject inference, where labels of most of the data are missing. However, we identified a gap in the RI literature, as no work has yet applied outlier detection in an iterative and controlled manner. This could help avoid biased assumptions that affect most extrapolation methods, which, despite their flaws, remain one of the most promising groups of RI techniques.

\subsection{Research hypotheses} \label{ssec:hypotheses}

The following hypotheses drive this study:

\begin{enumerate}
    \item The data from accepted clients alone is insufficient to train a credit scoring model that operates fairly across all potential applicants.
    \item The difference between distributions from accepted to rejected clients can be big; thus, extrapolating information from accepted to rejected should be done carefully.
    \item The behavior of some rejected clients can be inferred based on the data from accepted clients, allowing for a more comprehensive understanding of the applicant pool. Thus, to infer the rejected clients' behavior successfully, it is necessary to use distributional information about the clients.
    \item Evaluating reject inference (RI) models using only data from accepted clients fails to represent these models' true performance accurately.
\end{enumerate}

\section{Methodology}

\subsection{Proposed Framework}
\label{framework}

In this research, we propose a novel framework for RI that presents a semi-supervised learning method combining outlier detection (OD) and a confidence rule to infer the unlabeled sample classes. We call this framework Confident-Inline Extrapolation for Rejection Inference (CI-EX). Our approach, as many other studies involving RI and OD, is inspired by the methodologies of \cite{xia_novel_2019}. However, we do not use OD as a classification tool for the rejected data. Instead, we chose an approach similar to that of \cite{nikita_kozodoi_shallow_2019}, who also proposed an iterative method. However, unlike their approach, we did not use OD to filter out outliers but to select inlier samples at each iteration. And, differently from \cite{shih_framework_2022}, we propose an iterative method in which OD is not the actual labeler tool but only a filter step.

\begin{table*}[htbp]
\centering
caption{{Mathematical notations for Algorithm 1 and 2}}
  \label{tab:matnot}
  \begin{tabular}{cl}
    \toprule
    Notation & Description\\
    \midrule
    $X_{train}$ & set with labeled data\\
    $Y_{train}$ & set with labels\\
    $X_{rej}$ & set with unlabeled data\\
    $\eta$ & the number of samples to be added\\
    $\rho$ & ratio expected between good and bad payers\\
    $c$ & desired number of samples to be retrieved\\
    $\Delta$ & class (0 - non-default, 1 - default)\\
    \midrule
    $X_{\Theta}$ & set with inliner samples\\
    $X_{\Delta}$ & set with retrieved data\\
    $Y_{\Delta}$ & inferred labels for retrieved data\\
    $X_{train_\Delta}$ & set with data labeled as $\Delta$\\
    $x_j$ & feature vector of example j\\
    $P(X_{rej} = \Delta)$ & probability of $X_{rej}$ being $\Delta$\\
  \bottomrule
\end{tabular}
\end{table*}

To identify the samples from the rejected set we are most confident are from a specific class $\Delta$, we propose an algorithm that performs a two-step verification on each sample. First, we check if the rejected sample is a non-outlier for the class $\Delta$. Then, we check if that sample belongs to the subset of $c$ samples with the highest probability of belonging to that class. At each iteration, our algorithm labels and adds $\eta$ samples from the unlabeled set to the training set and removes those samples from the unlabeled set.

\subsubsection{Retrieve Confident Samples}

The Retrieve Confident Samples Algorithm ({\Cref{alg:retrieveconfident}}) describes the core of our framework, and {\Cref{tab:matnot}} constitutes a quick guide for better reading of our proposed algorithms. The algorithm takes as input a labeled training dataset, where $X_{train}$ represents the informative features of the dataset, and $Y_{train}$ represents the target feature of the dataset. Due to changes during method iterations, $Y_{train}$ consists of both ground truth labels and inferred labels, and $X_{train}$ may consist of both accepted and rejected client data. The proportion between accepted and rejected data will depend on the current iteration of the framework as new data is added to the training set.

As mentioned before, our framework employs a two-step verification to ensure that the rejected samples added at each iteration are more likely to be the ones we are most confident will get the inferred labels. The first step uses Isolation Forest \cite{liu2008isolation}, an outlier detection algorithm, to divide the rejected samples between outliers and non-outliers. Instead of fitting the model with the entire training set, we fit the model with one class at a time from the training set ($X_{train_\Delta}$). Our first hypothesis is that samples considered non-outliers based on $X_{train_\Delta}$ are likelier to belong to that class. In \Cref{alg:retrieveconfident}, this set of samples considered non-outliers, $X_{\Theta}$, moves on to the next step. We have experimented with two modes for labeling the rejected set, which will be described subsequently.

\subsubsection{Extrapolation Mode}

\begin{figure}[htb]
    \centering
    \includegraphics[width=1\linewidth]{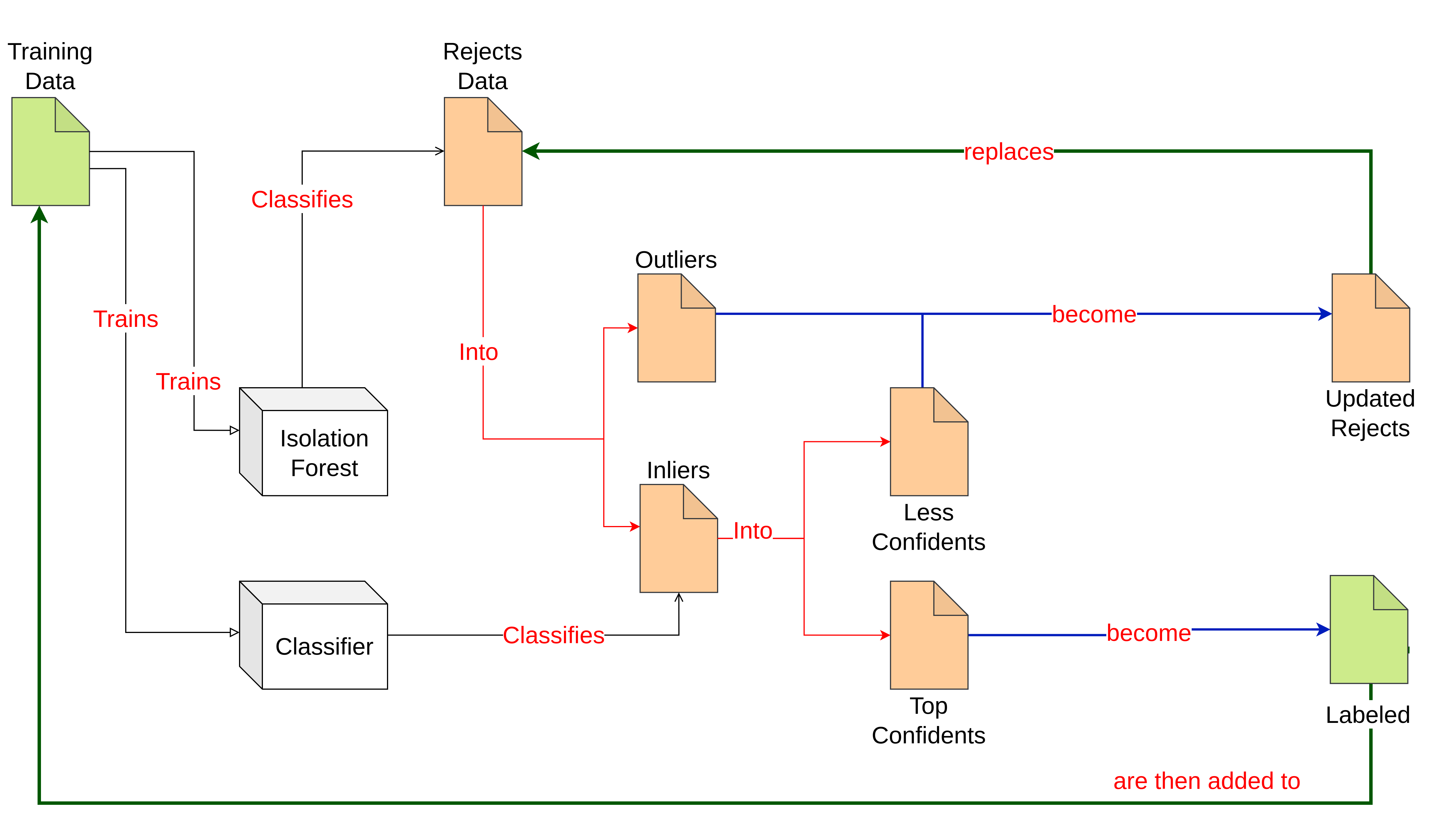}
    \caption{Representation of CI-EX framework to perform Reject Inference.}
    \label{fig:CI-EX}
\end{figure}

We call this version of the proposed framework Confident-Inline Extrapolation (CI-EX). \Cref{fig:CI-EX}, illustrate how this version of \Cref{alg:retrieveconfident} works. As can be seen in the figure and in steps 2 and 3(a) of the algorithm, in the CI-EX mode, the current training data is used to train the Isolation Forest algorithm and the classifier. However, although the whole training data is used to train the classifier, only the samples from the respective class $\Delta$ are used to train the Isolation Forest. Because of this, the \Cref{alg:retrieveconfident} needs to be executed twice at each iteration, returning $c$ samples with label $\Delta$ --- or fewer if fewer than the stipulated number of samples match the criteria. In step 4, using the Isolation Forest, we classify the Rejects Data into outliers and inliers. Outliers are ignored temporarily, but will belong to the updated rejects dataset at the end of the iteration of the \Cref{alg:retrieveconfident}. Inliers, however, are further subdivided into top confident, which go to set $X_{\Theta}$, and less confident samples.

\begin{algorithm}[ht]
\caption{{Retrieve Confident Samples}}
\label{alg:retrieveconfident}
\begin{algorithmic}[1]
\State \textbf{Input:} $X_{train}, Y_{train}, X_{rej}, \Delta, c$
\State \textbf{Output:} $X_{\Delta}, Y_{\Delta}$
\State $X_{\Delta} \gets \{\}$, $Y_{\Delta} \gets \{\}$, $N \gets |X_{rej}|$

\State Fit IsolationForest with $X_{train_{\Delta}}$
\State Fit Classifier with $X_{train}, Y_{train}$

\State $X_{\Theta} \gets \{ x_i \in X_{rej} \mid \text{outlier}(x_i) = \text{False}, i=1,\dots,N \}$

\While{$|X_{\Delta}| < c$ and $|X_{rej}| > 0$}
    \State $x_j \gets \operatorname*{argmax}_j P(X_{rej} = \Delta)$
    
    \If{$\text{score}(x_j) \geq 0.5$}
        \State $y_j \gets 1$
    \Else
        \State $y_j \gets 0$
    \EndIf
    
    \If{$x_j \in X_{\Theta}$}
        \State $X_{\Delta} \gets X_{\Delta} \cup \{x_j\}$
        \State $Y_{\Delta} \gets Y_{\Delta} \cup \{y_j\}$
    \EndIf
    
    \State $X_{rej} \gets X_{rej} - \{x_j\}$
\EndWhile
\end{algorithmic}
\end{algorithm}

With this strategy, step 5.1 of \Cref{alg:retrieveconfident} uses the probabilities derived from a classifier with balanced weights\footnote{In our implementation, instead of using the default learning procedure that makes all samples equally important, the weight of each sample is inversely proportional to the number of samples of its class.} to label the inliers samples and filter the $c$ most confident samples\footnote{Since our classifier uses balanced weights, we can use $0.5$ as the threshold to classify the samples between \textit{good} and \textit{bad} cases.} (steps 5.2 to 5.3).  However, the less confident samples will also become part of the updated rejects dataset at step 5.4  of the iteration of the \Cref{alg:retrieveconfident}.  The algorithm then returns $c$ samples with label $\Delta$ --- or fewer if fewer than the stipulated number of samples match the criteria. After the \Cref{alg:retrieveconfident} is executed for both classes, the rejected and training datasets are updated, as illustrated in \Cref{fig:CI-EX}.

\subsubsection{Expand Dataset}

The Expand Dataset Algorithm {(\Cref{alg:expanddataset})} can be understood as an iteration of our framework. We take as input a labeled train set ($X_{train}$ and $Y_{train}$) and an unlabeled set ($X_{rej}$), and two other parameters, $\eta$ and $\rho$, to control how many good and bad cases should be added to the training set at this iteration. The parameter $\eta$ is the total number of samples we want to add at this iteration, and the parameter $\rho$ defines the proportion of bad to good payers in the total number of added samples we want to add. 

We then call the Retrieve Confident Samples Algorithm {(\Cref{alg:retrieveconfident})} within the Expand Dataset Algorithm for both classes, to retrieve $c_0$ samples with inferred labels for the class \textit{good payers} ($X_{\Delta=0}$, $Y_{\Delta=0}$), and $c_1$ samples with inferred labels for the class \textit{bad payers} ($X_{\Delta=1}$, $Y_{\Delta=1}$). The samples inferred for both groups are then concatenated to the new training set ($\hat{X}_{train}$ and $\hat{Y}_{train})$ and removed from the unlabeled set, $X_{rej}$. The expanded training and updated unlabeled sets are returned as the algorithm output.

\begin{algorithm}[ht]
\caption{{Expand Dataset}}
\label{alg:expanddataset}
\begin{algorithmic}[1]
\State \textbf{Input:} $X_{train}, Y_{train}, X_{rej}, \eta, \rho$
\State \textbf{Output:} $\hat{X}_{train}, \hat{Y}_{train}, \hat{X}_{rej}$
\State $c_{0} \gets \eta - (\eta \cdot \rho)$
\State $c_{1} \gets \eta \cdot \rho$

\State $(X_{\Delta=0}, Y_{\Delta=0}) \gets \text{RetrieveTS}(X_{train}, Y_{train}, X_{rej}, 0, c_{0})$
\State $(X_{\Delta=1}, Y_{\Delta=1}) \gets \text{RetrieveTS}(X_{train}, Y_{train}, X_{rej}, 1, c_{1})$

\State $\hat{X}_{train} \gets \text{Concat}(X_{train}, X_{\Delta=0}, X_{\Delta=1})$
\State $\hat{Y}_{train} \gets \text{Concat}(Y_{train}, Y_{\Delta=0}, Y_{\Delta=1})$
\State $\hat{X}_{rej} \gets X_{rej} \setminus \hat{X}_{train}$
\end{algorithmic}
\end{algorithm}

\subsection{Data}
This research uses data from the HomeCredit European dataset~\cite{homecredit}. As well as from the Lending Club dataset~\cite{lendingclub}, a popular online credit loan platform in the US \cite{Liu2022}, and used for much research in credit scoring. They are two of the most extensive credit datasets publicly available online. Both datasets were made available on the Kaggle website\footnote{https://www.kaggle.com/}, where competitions related to the identification of bad payers in credit scoring scenarios using these datasets were held.

For the Homecredit dataset, from different files with varying levels of information about the client's data, we consider only the information present in the \textit{application\_{train}.csv} file for this study. It contains $307,507$ samples from approved clients, with 122 informative features and one target feature. Of the informative features, 106 were numerical, and 21 were categorical. Other files were not considered for this study since they were composed chiefly of information that would only be available for approved clients. This data type would not be helpful for us, as we focus only on the credit-granting process.

The Lending Club dataset contains an even more extensive amount of credit data: 2260701 samples for accepted clients and 27648741 samples from rejected clients from 2007 to 2018. Due to this, it can be utilized to train and test a reject inference credit scoring model sufficiently well. This dataset comprises tabular data and contains 151 features for the accepted clients but only nine for the rejected clients. Data from accepted clients can be labeled between \textit{good} and \textit{bad} debtors using the column \textit{Loan\_Class}. This data was used to train and test our supervised models. The rejected clients' data is unlabeled and was used to perform Rejected Inference.

\subsection{Data pre-processing}
\begin{figure}[htbp]
  \centering
    \includegraphics[width=1\linewidth]{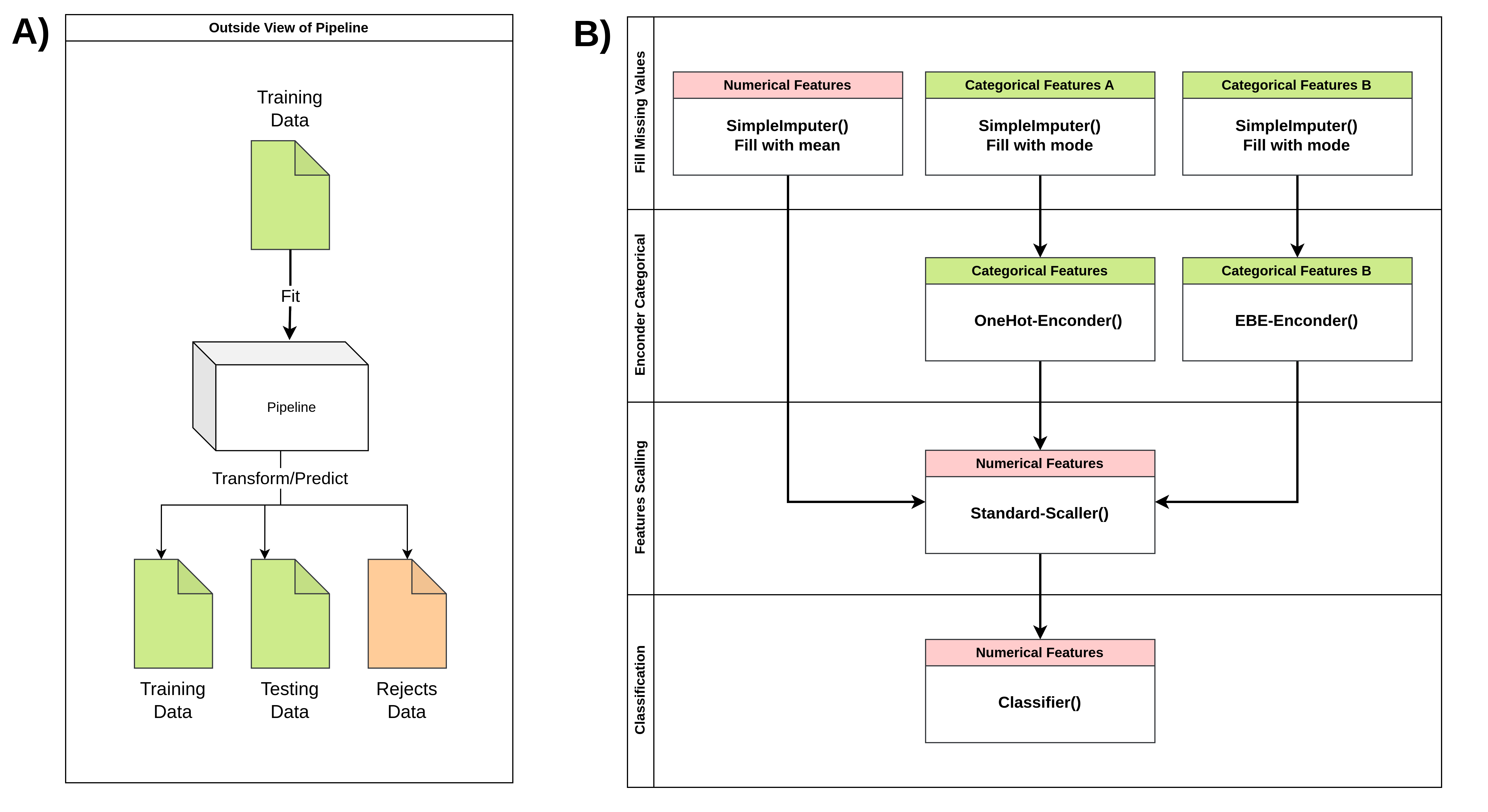}
  \caption{(A) Outside view of the pipeline. (B) Inside view of the Pipeline. The pipeline is fitted with the training set and used for pre-processing and classification on all datasets.}
  \label{fig:sklearn-pipeline}
\end{figure}

Most data pre-processing was made automatically using scikit-learn pi\-pelines~\cite{scikit-learn}. Due to their structure, which combines several steps of data pre-processing and classification, they are a helpful tool for data science. By fitting all models inside the pipeline, from processing to classification, with the training data, they also help to avoid data leakage from the testing set. As illustrated in the \Cref{fig:sklearn-pipeline} (A), the training data is used to fit the pipeline, and from that, the pipeline can be used to transform and make predictions. When a function is called to predict a testing set, it will apply transformations (pre-processing) to the testing set as necessary based on the values fitted with the training set.

The \Cref{fig:sklearn-pipeline} (B) describes the steps implemented on the pipeline created for our experiments. Our pipeline separates features into three categories: numerical Features, categorical Features A, and categorical features B. Group A has categorical features with less than $3$ unique values, and group B has at least $3$ unique values. Both groups of categorical features are submitted to the same type of null value filling. The mode of the feature is fitted and used to fill the possible missing values of that feature. The null values of the numerical features are filled with the mean instead. Group A is encoded with one-hot encoding, which replaces each categorical feature with a column for each unique value. The column will contain 1 if the sample has that unique value and 0 if it does not. Group B uses a more complex encoding based on Empirical Bayesian Estimation (EBE), available at the scikit-learn library as a Target Encoder. The Target Encoder replaces categorical values with a value that reflects the proportion of positive cases observed for each category during the fitting process. 

\subsubsection{HomeCredit}
We separated the informative features of the HomeCredit dataset into three subgroups $S_1$, $S_2$, and $S_3$. In $S_1$, we allocated the features we considered more relevant to the study of Reject Inference, such as information such as age, number of children, education, and score of a client in other sources, among others, totaling 15 informative features. $S_1$ features descriptions are listed in \Cref{table:hcdesc}\footnote{The descriptions are provided by the Kaggle repository.}. The $S_2$ subgroup consisted of 71 informative features such as the client's housing situation, the number of times the client's credit information was checked before the loan, and statistics about the building where the client lives. Finally, the $S_3$ subgroup of features was composed of features like sensitive information, such as gender, occupation, and family status of the client, as well as extremely unbalanced features like binary features with information about certain documents, where more than $99\%$ of values were the same for all samples.

For the Lending Club dataset, we took inspiration from the work of \cite{shih_framework_2022} to make our feature selection for the accepted and rejected clients dataset. However, we decided to avoid certain features in the dataset that would lead to target leaking. These were features related to the credit payment behavior of the client and, thus, were not available to the rejected population. The feature descriptions for the accepted client's dataset are available at \Cref{table:lcdesc-acp}. Respectively, \Cref{table:lcdesc-rej} brings the descriptions of the selected features for the rejected clients. The \textit{issue\_d} feature on \Cref{table:lcdesc-acp} and \textit{Application Date} on \Cref{table:lcdesc-rej} feature were used to separate the datasets between train and test and were not used to train the models.

\subsection{Evaluation metrics}
\subsubsection{Area Under the Curve}

One evaluation problem in credit scoring is the class imbalance in credit risk datasets. To bypass this problem, metrics such as Area Under the Curve (AUC) are welcomed. AUC is a metric that is not sensitive to threshold values. The higher the AUC value, the better the classifier \cite{Dastile2021}. It also reflects the model's performance, even when dealing with unbalanced datasets. The {\bf Area Under the Curve} is given by:
\begin{equation}
AUC = P[p(y=1|X_i)>p(y=1|X_j)|y_i=1, y_j=0]    
\end{equation}

\subsubsection{Kickout}

\begin{figure}
    \centering
    \includegraphics[width=0.85\linewidth]{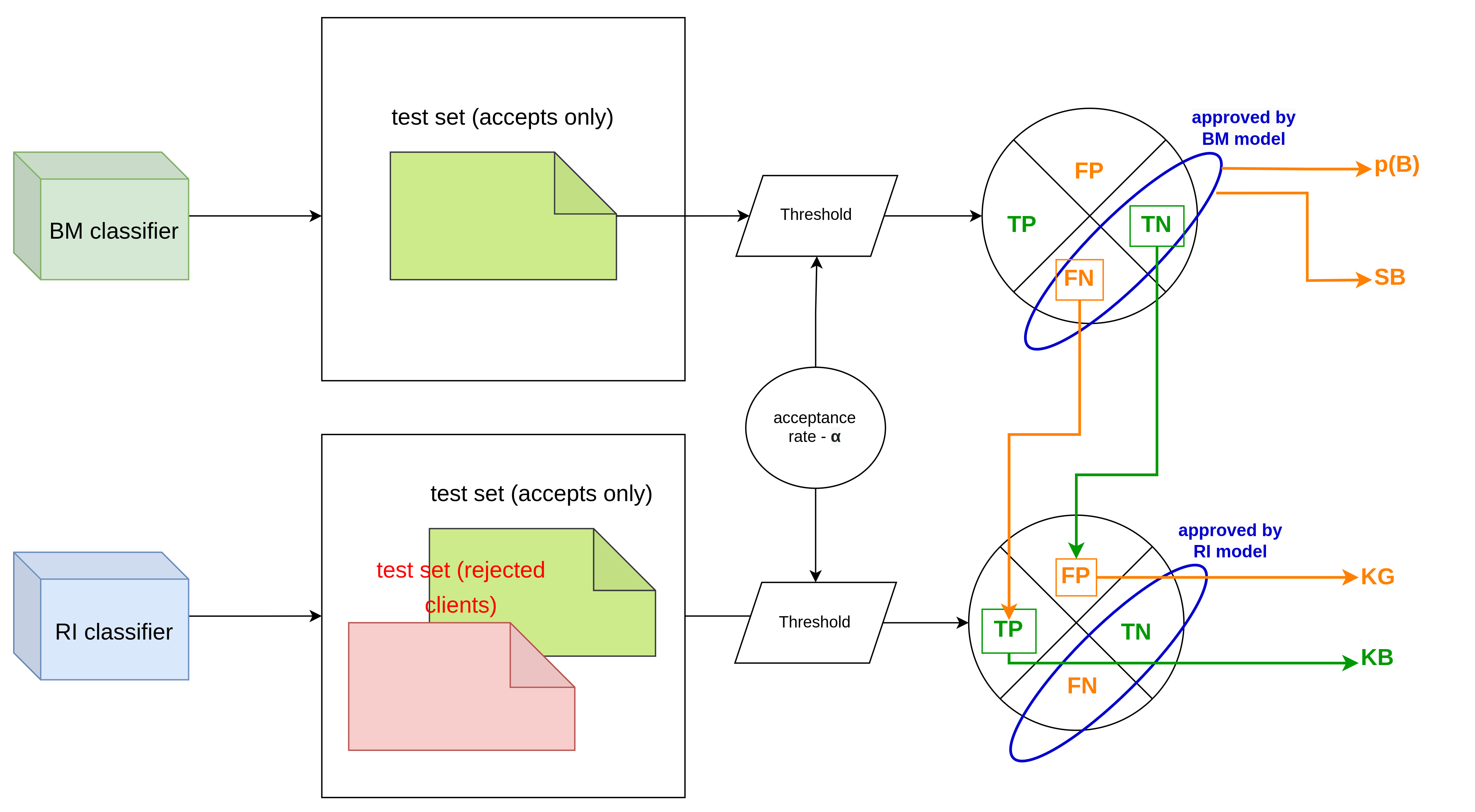}
    \caption{Illustration of how the Kickout metric is calculated. TP stands for True Positives, FP for False Positives, TN for True Negatives, and FN for False Negatives.}
    \label{fig:kickout-metric}
\end{figure}

Kickout, proposed by \cite{nikita_kozodoi_shallow_2019}, is a metric that aims to evaluate the performance of a RI model relative to a benchmark model. As illustrated on \Cref{fig:kickout-metric}, to calculate this metric, we need a labeled test set (from the accepts) and an unlabelled test set (from the rejects). The labeled test set is used to evaluate the benchmark model, and both datasets are used to assess the RI model. It evaluates the number of good and bad cases the model accepts with and without using RI, following the formula in \Cref{eq:kickout}. Considering that we have a benchmark model (BM) without RI, with $S_B$ bad payers, $K_B$ is the number of bad payers accepted by the benchmark model (i.e. false negative cases) now rejected by a method with RI, and $K_G$ is the number of good payers accepted in the benchmark model (i.e. true negative cases) now rejected by a technique with RI. $p(B)$ and $1-p(B)$ are the probabilities of bad and good payers, given that the benchmark model has accepted them. So, $\frac{K_B}{p(B)} - \frac{K_G}{1-p(B)}$ is the difference in the numbers of bad to good payers in proportion to the number of bad and good payers accepted by the benchmark model. And $\frac{S_B}{p(B)}$ is the ratio between the number of ground truth bad payers accepted by the benchmark model and the probability of a ground truth bad payer being accepted by the benchmark model. A good RI model is expected to have a higher kickout value. 

\begin{equation}
    \text{kickout} = \frac{\frac{K_B}{p(B)} - \frac{K_G}{1-p(B)}}{\frac{S_B}{p(B)}}
    \label{eq:kickout}
\end{equation}

This metric is essential in credit scoring because it can capture the risk of giving credit to bad payers when we include more clients using reject inference. The acceptance rate, $\alpha$, defines the proportion of clients the models will accept. The decision threshold that separates the clients between accepted and rejected is calculated in the accepted set for the benchmark and in the accepted and rejected set using RI. In the last case, we give credit to more people, and the kickout evaluates how well our exclusion of bad payers went in the RI scenario. A higher kickout reflects a better quality score system when compared with the benchmark.

The importance of the acceptance rate, $\alpha$, is worth mentioning. This parameter can create different kickout values depending on its selection. \cite{nikita_kozodoi_shallow_2019} did not explicitly stipulate any value to this variable. For these reasons, we also made a study that evaluated all RI techniques by a range of values for $\alpha$.

\subsubsection{Area Under the Kickout}

Our study of how different values of $\alpha$ create an enormous range of kickout values. This leads us to realize that focusing on a single value for $\alpha$ may lead to biased conclusions. Therefore, we propose a new metric called Area Under the Kickout (AUK). This metric evaluates the mean of the kickout values for each value of $\alpha$ ranging from 1\% to 100\%. The formula for calculating the AUK value is given by \Cref{eq:auk}. In the equation, $\alpha$ represents the percentage of clients the model accepts. The bigger the value of the AUK, the better the model identifies bad clients.

\begin{equation}
    \text{AUK} = \dfrac{\sum_{\alpha=1}^{100}\text{kickout}(\alpha)}{100}
    \label{eq:auk}
\end{equation}

\subsection{Parameter Selection and Experimental Setup}

{The design of our proposed algorithm for selecting the most confident samples is governed by two primary hyperparameters: the number of samples added to the updated training set at each iteration ($\eta$) and the proportion of relabeled positive-class samples incorporated into the updated training set ($\pi$). We conducted experiments to examine the correlation between $\eta$, $\pi$, and our metrics of interest, with detailed results provided in \Cref{table:combinations_table}. The table outlines the random seeds, $\pi$, and $\eta$ values analyzed. Each $\pi$ and $\eta$ combination was tested across five random seeds, resulting in $275$ experiments.}

\begin{table}[ht]
\centering
\small
\caption{Combinations of random seeds, proportion of relabeled positive-class samples (\(\pi\)), and sample sizes (\(\eta\)) used in experiments to analyze the influence of hyperparameters on AUC and AUK metrics.
}
\begin{tabular}{p{0.10\textwidth}p{0.15\textwidth}p{0.1\textwidth}}
\hline
\textbf{Seeds}         & \textbf{Percent Bads ($\pi$)} & \textbf{Size ($\eta$)} \\ \hline
120054, 388388, 570334, 907360, 938870 & 0.07, 0.08, 0.09, 0.1, 0.12, 0.13, 0.14, 0.15, 0.16, 0.18, 0.2, 0.22, 0.24, 0.26, 0.28, 0.3, 0.32, 0.36, 0.4 & 1000, 5000, 10000
\end{tabular}
\label{table:combinations_table}
\end{table}

\begin{figure}[htb]
  \centering
    \includegraphics[width=0.9\linewidth]{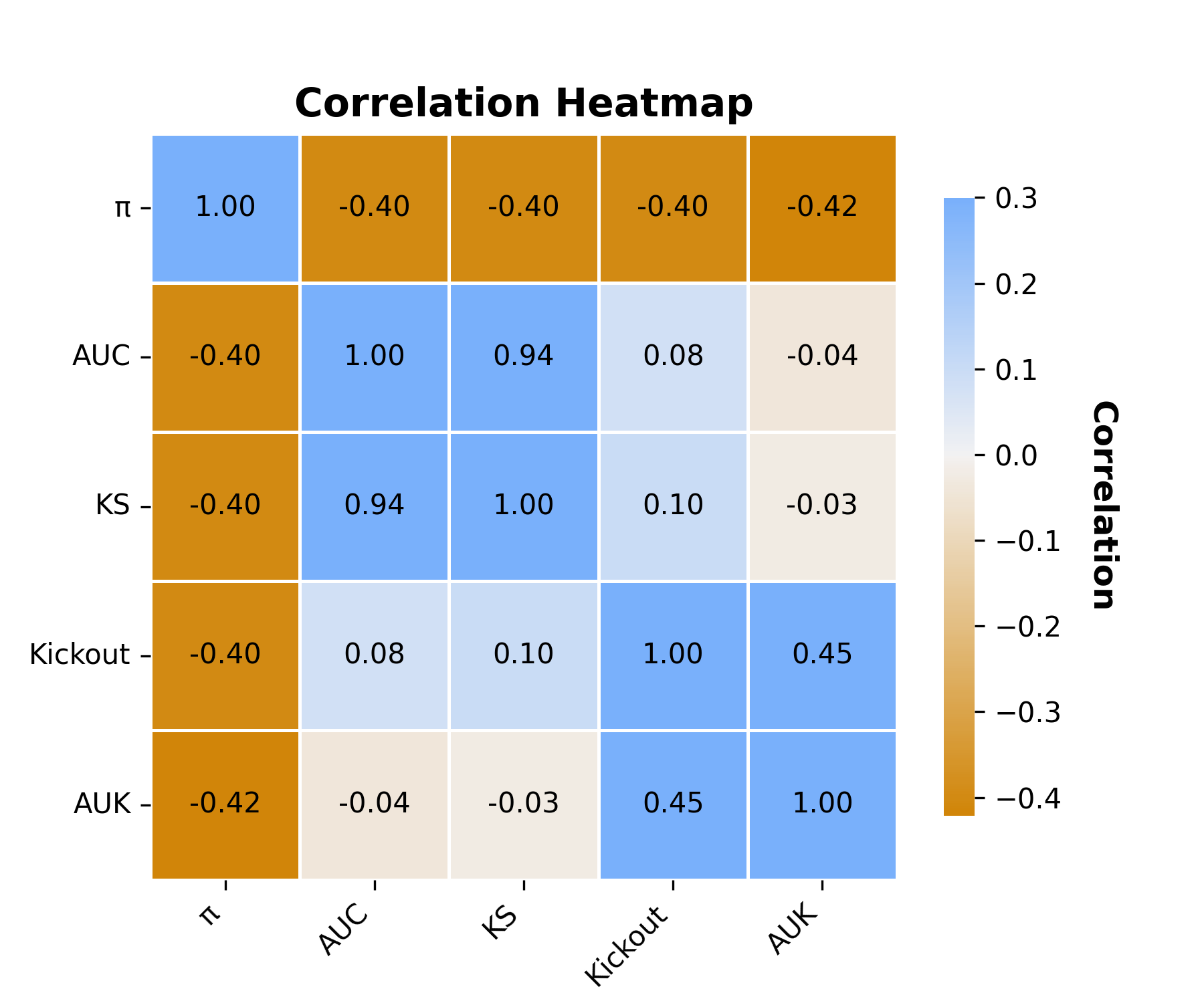}
  \caption{Correlation of $\pi$ value and the metrics studied.}
  \label{fig:corr_val_2010}
\end{figure}

\begin{figure}[htb]
  \centering

\begin{subfigure}[b]{0.49\textwidth}
    \centering
  \centering
    \includegraphics[width=\linewidth]{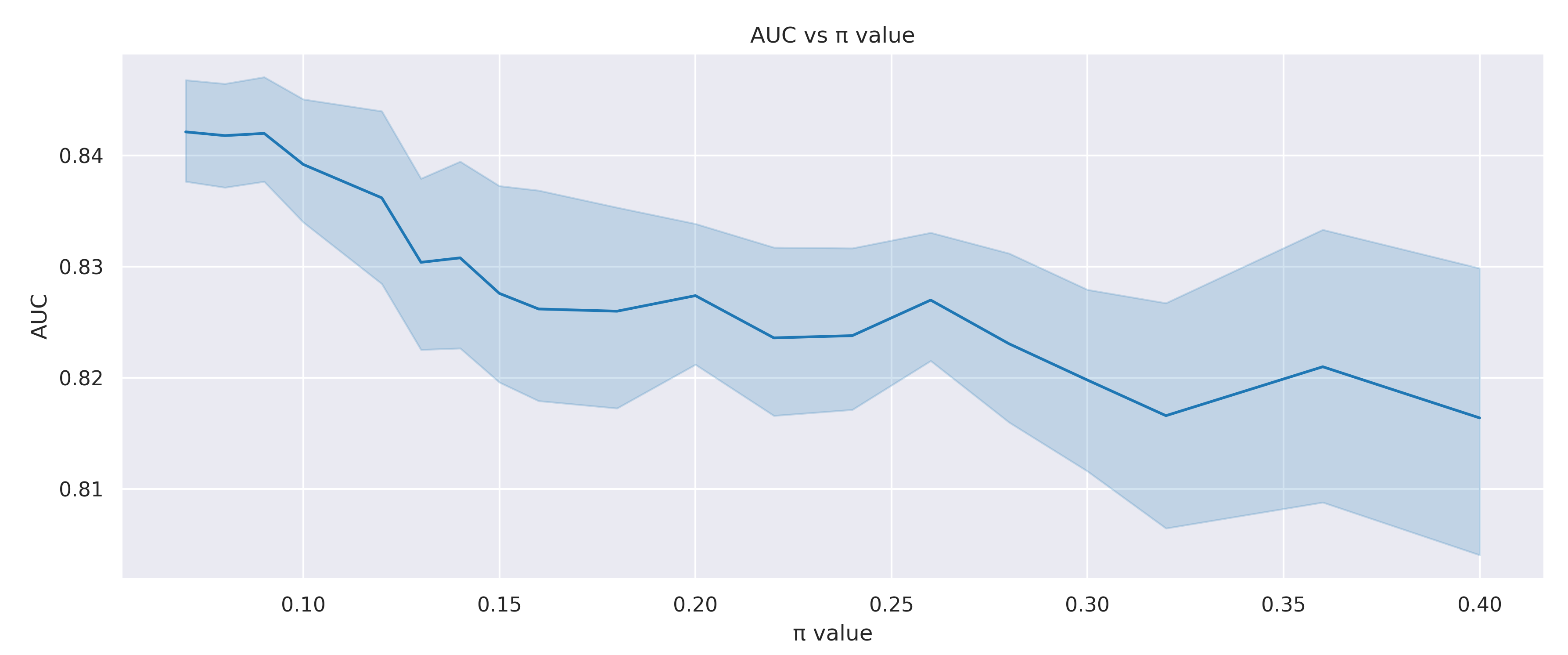}
\caption{AUC}
  \label{fig:auc_pi}
    \end{subfigure}
\begin{subfigure}[b]{0.49\textwidth}
    \centering
    \includegraphics[width=\linewidth]{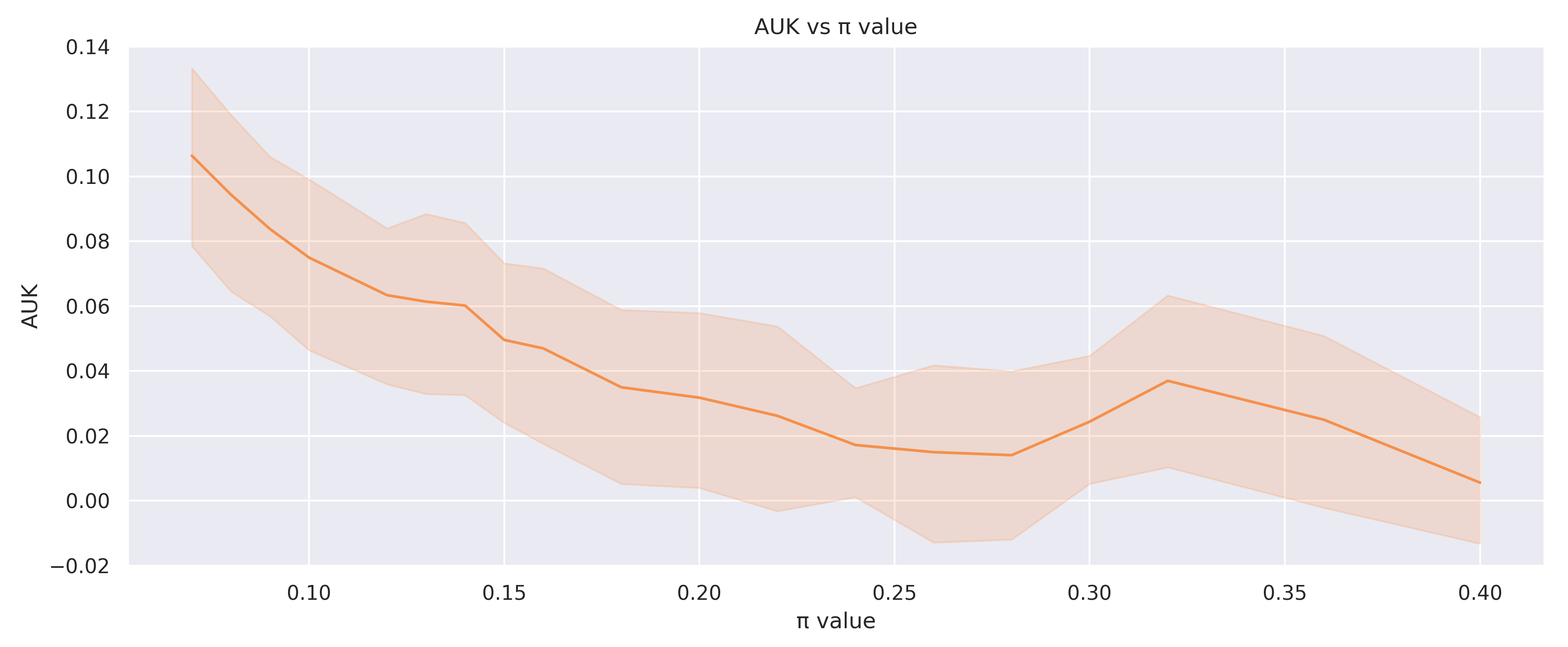}
    \caption{AUK}
  \label{fig:auk_pi}
    \end{subfigure}
\caption{Relationship between the AUC (a) and AUK (b) metric and $\pi$ values. The plot presents the average AUC (a) or AUK (b) value, along with the 95\% confidence interval, as a function of different $\pi$ values, highlighting the variation in performance as $\pi$ changes.}
\end{figure}

\Cref{fig:auc_pi} and \Cref{fig:auk_pi} illustrate the influence of $\pi$ on the final AUC and AUK metrics, respectively. Both figures show a clear downward trend in AUC and AUK values as $\pi$ increases, indicating a strong negative correlation between $\pi$ and these metrics. This observation is further supported by \Cref{fig:corr_val_2010}, which presents the correlation matrix, showing a slight negative correlation between AUC and AUK, attributed to expected trade-offs in dataset debiasing. Other experiments showed that variations in $\eta$ showed no significant correlation with the metric values (See \ref{app1}).

\subsubsection{Model Selection and Classifier Optimization}

We selected LightGBM \cite{ke2017lightgbm} as our primary classifier, given its high performance and training efficiency with tabular data, particularly with the HomeCredit dataset as verified by \cite{daoud_comparison_2019}. Implementations of the Label Spreading Algorithm and Isolation Forest were sourced from the Sklearn library in Python \cite{scikit-learn}. Hyper-parameter optimization for LightGBM was conducted on the accepted training and validation sets, with the resulting optimized parameters consistently applied across all instances where LightGBM was used. Each RI technique utilized the same set of optimized parameters, eliminating the need for further tuning and ensuring a fair comparison across techniques. It is worth mentioning that despite using LightGBM as the classifier, the proposal is model-agnostic.

\subsubsection{Dataset Selection Using TOPSIS}

{Both of our techniques generate progressively larger training datasets. A classifier trained on each dataset is evaluated in two metrics, AUC and Kickout (for a specific acceptance rate $\alpha$). We selected the optimal dataset and classifier using the multi-criteria decision-making TOPSIS method \cite{chakraborty_topsis_2022}. This approach allowed us to identify a dataset that provides a good balance between AUC, weighted at \(1\), and kickout value, weighted at \(10\), based on the specified \(\alpha\) value. A higher weight was assigned to kickout, which is considered a more relevant metric in this context. Preliminary experiments on the validation set indicated that prioritizing kickout resulted in only a slight reduction in AUC. In this study, we set \(\eta\) to \(1000\), \(\rho\) to \(0.07\), and the contamination threshold for the Isolation Forest algorithm to \(0.12\), with these parameters obtained through manual fine-tuning.}

\subsection{Experimental design}

\subsubsection{Experiment I}

{In Reject Inference, two types of datasets are necessary --- labeled data from the accepted population and unlabeled data from the rejected population. It is essential to compare how employing rejected clients' information will improve the credit scoring system concerning the benchmark model. However, finding public datasets with rejected samples can be pretty challenging \cite{liu2022rmt}. One way to surpass such a limitation is simulating rejected clients using accepted-only datasets, as done by \cite{liu2022rmt}. Therefore, we simulated different accept/reject policies using the HomeCredit accepted clients' data to access both accepted and rejected data distributions in this experiment. We used this data to evaluate the different RI techniques studied in this paper and the one we proposed. In this experiment, we aimed to verify how simulated rejected data can be applied to validate RI methodologies. }

\begin{figure}[htbp]
  \centering
    \includegraphics[width=0.7\linewidth]{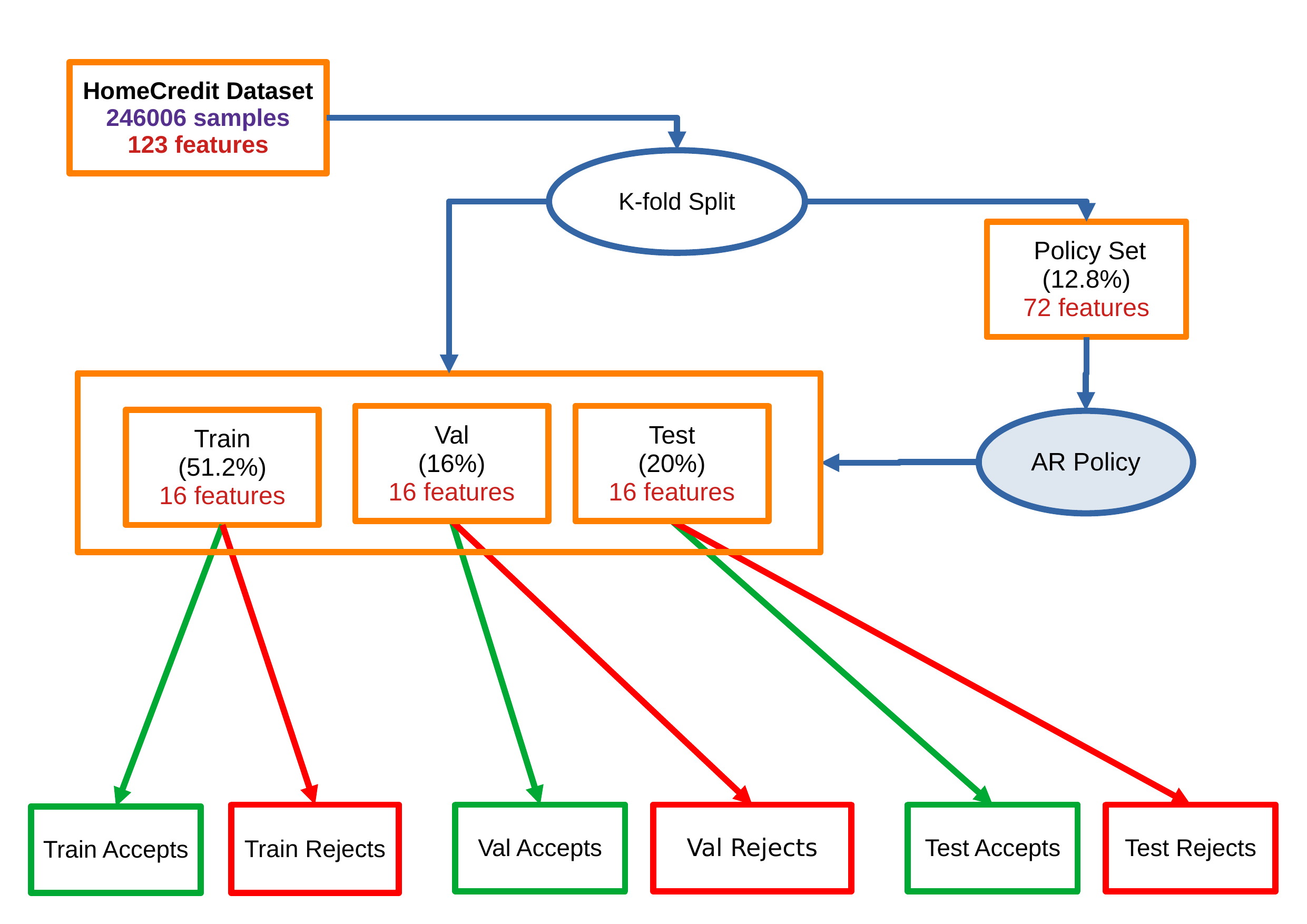}
  \caption{The split of the HomeCredit dataset into seven subsets}
  \label{fig:accept-reject-policy}
\end{figure}

{ \Cref{fig:accept-reject-policy} outlines our methodology for splitting the datasets into different subsets of accepted and rejected clients. Each \textit{Rejects} subset would be considered unlabeled data in this methodology. Generating each subset starts with isolating $20\%$ of samples from the dataset and the cherry-picked features to fit a Logistic Regression classifier. The use of Logistic Regression here is inspired by the work of \cite{nikita_kozodoi_shallow_2019}. According to the authors, this weak learner with L1 regularization is a more reliable way to use the probabilities of default given by the model as a separator between the two classes.  In this simulation, we define $\epsilon$ as the threshold value that distinguishes good clients from risky ones. Any sample assigned a probability of default greater than $\epsilon$ is categorized into the rejected group. Experiments were conducted using $\epsilon$ values within the range of $[0.3, 0.65]$. }

\begin{figure}[htbp]
  \centering
    \includegraphics[width=0.9\linewidth]{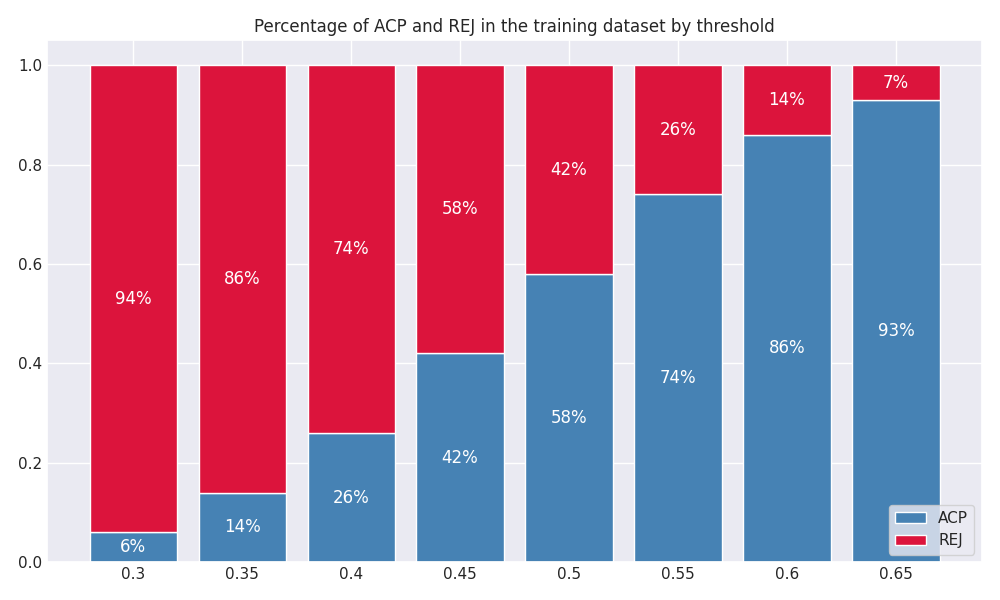}
\caption{The split of the training set of the HomeCredit dataset by threshold value. }
  \label{fig:accept-reject-split}
\end{figure}

{\Cref{fig:accept-reject-split} illustrates the resulting proportions of accepted and rejected clients based on the $\epsilon$ values in the training set. In real-world scenarios, the number of rejected clients typically exceeds the number of accepted ones, making the training sets generated with $\epsilon$ values below 0.5 the most realistic \cite{nikita_kozodoi_shallow_2019, liu2022rmt}. Therefore, we generated seven distinct subsets for each experiment run, as shown in \Cref{table:dataset-categories} (See also \ref{tab:number_accept_reject_split}). The simulated accept/reject policy was fitted with 20\% of the initial accepted set --- these samples were ignored until the next run. Each configuration was run 20 times with a different random seed to ensure robustness.}

\begin{table*}[htbp]
\centering
  \caption{{Description of Dataset Categories}}
  \label{table:dataset-categories}
  \begin{tabular}{clll}
    \toprule
    Category & Description \\
    \midrule
    Policy Set & The set used only for fitting the accept/reject policy  \\
    Train Accepts & Labeled training set\\
    Train Rejects & Unlabeled training set\\
    Val Accepts & Set used to evaluate the best iteration of our method \\
    Val Rejects & Set used to evaluate the best iteration of our method\\
    Test Accepts & Set used to evaluate all methods\\
    Test Rejects & Set used to evaluate the kickout Metric \\
    \bottomrule
  \end{tabular}
\end{table*}

\subsubsection{Experiment II }

\begin{figure}[h]
  \centering
    \includegraphics[width=0.7\linewidth]{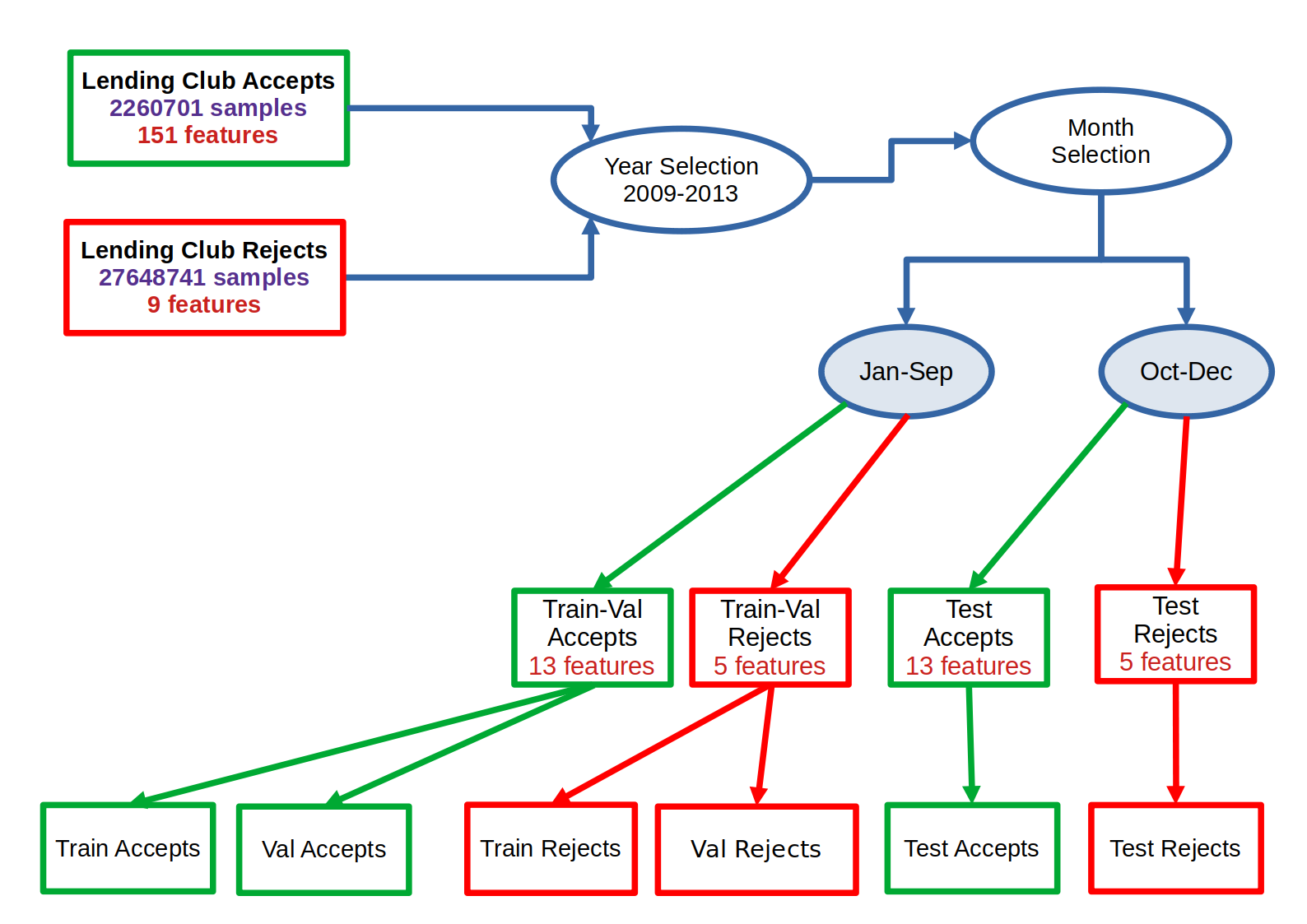}
  \caption{{The split of the Lending Club dataset into six subsets}}
 
  \label{fig:LC-policy}
\end{figure}

In this experiment, we aimed to verify whether our proposed framework could correctly identify bad payers. We chose a different approach for the experiments using the Lending Club dataset. Instead of random K-fold validation, we separate the training and testing sets by time. So, for each specific year, the training set is composed of the loans dated from January to September, and the testing set is composed of the loans dated from October to December. However, the training and validation sets were created using the widely adopted train and test split function from the Scikit-learn library. $70\%$ of the initial training set was kept as training, and the remaining $30\%$ was used as validation. We followed this protocol for both the accepted and rejected clients' datasets. Ultimately, we got six distinct subsets for each year studied. Each subset was used for the purposes listed in \Cref{table:dataset-categories}. \Cref{fig:LC-policy} describes the data separation protocol utilized. 

\Cref{fig:LC-policy} also lists the years chosen to be analyzed in this research, 2009 to 2012, and the final number of features selected from each dataset. We took inspiration from the work of \cite{shih_framework_2022} to select the feature. We also based our year selection on their work. However, the bigger inspiration from their work was the data imputation on missing features. As shown in \Cref{fig:LC-policy}, there is a different disparity in the number of features for the rejected and accepted client datasets, which is a big obstacle in machine learning research. \cite{shih_framework_2022} overcomes this issue by applying the k-nearest imputation method to fill out the values of missing features on the rejected dataset. This method was proposed by \cite{troyanskaya2001missing}, and works by using sample similarity between the accepted training dataset and the rejected datasets to estimate the missing values of the features on the latter, based on the former.

In this experiment, for our proposed technique, we set $\eta$ as $1000$, $\rho$ as $0.2$, and $0.12$ as the value for the contamination threshold for the Isolation Forest algorithm. The value of these parameters was obtained through manual fine-tuning. We also chose the TOPSIS method for this experiment to identify the best version from the resulting datasets obtained by our proposed techniques. However, instead of selecting kick-out values, we used the AUK metric, which is less biased toward an acceptance rate value.

\section{Results}

{The \Cref{fig:pca_2010} presents the evolution of the dataset through PCA across several iterations of the algorithm. Initially, the dataset consists of accepted and rejected clients; as stated in the hypotheses in Section \ref{ssec:hypotheses}, the distributions are different. As the algorithm progresses, several samples from the rejected dataset are gradually added to the accepted dataset at each iteration. This incremental incorporation of previously rejected clients results in a shift in the distribution of the accepted dataset. This procedure helps us to find the right moment where adding more samples can hurt the performance of the credit scorer, following the hypotheses in Section \ref{ssec:hypotheses}.

{Throughout iterations, it can be observed that the accepted dataset undergoes a gradual expansion, reflecting the inclusion of more diverse clients. The PCA visualization highlights how the distribution of accepted clients becomes broader and more encompassing as the dataset evolves to include a larger and more varied portion of the rejected clients. This progression changes the dataset’s structure and signifies a key contribution of the algorithm, allowing the model to better generalize by adapting the training data to be more representative of the overall client base. }

\begin{figure}[htbp]
    \centering
    \includegraphics[width=1\linewidth]{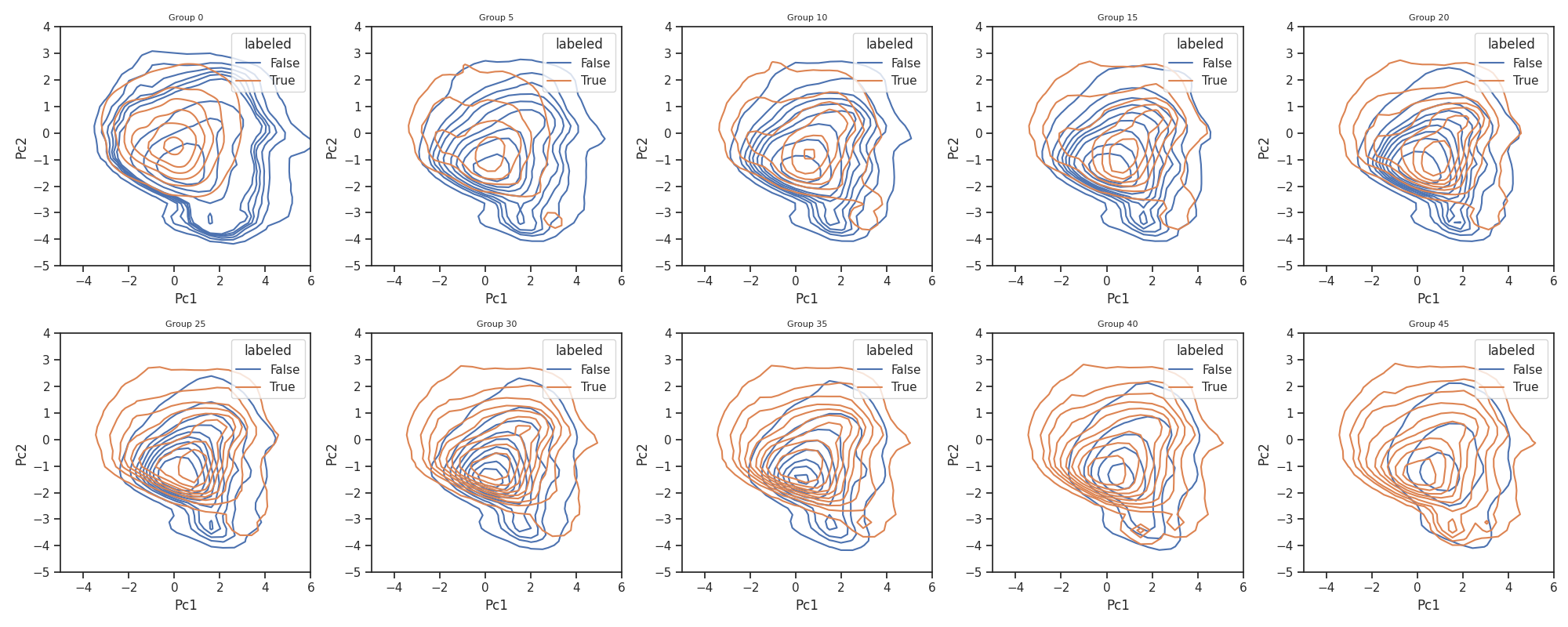}
    \caption{Kernel Density Estimation (KDE) plot after applying Principal Component Analysis (PCA) on the dataset Lending Club (year 2010). Parameters: $\eta = 1000$ and $\pi=0.07$}
    \label{fig:pca_2010}
\end{figure}

\subsection{Results Using Real Rejected Clients}

{Given the high number of samples available for the Lending Club dataset, we decided to separate each year as a sub-experiment, following the experimental design mentioned in the previous section. This way, we could experiment not only with different dataset sizes but also with different distributions and periods. We analyze our results using the AUC, in \Cref{fig:auc-heatmap-lc}, and AUK, in \Cref{fig:auk-heatmap-lc}, metrics. We also analyze how these metrics relate in a multi-objective perspective presented in \Cref{fig:auc-auk-lc}. } 

\begin{figure}
        \centering
        \includegraphics[width=1\linewidth]{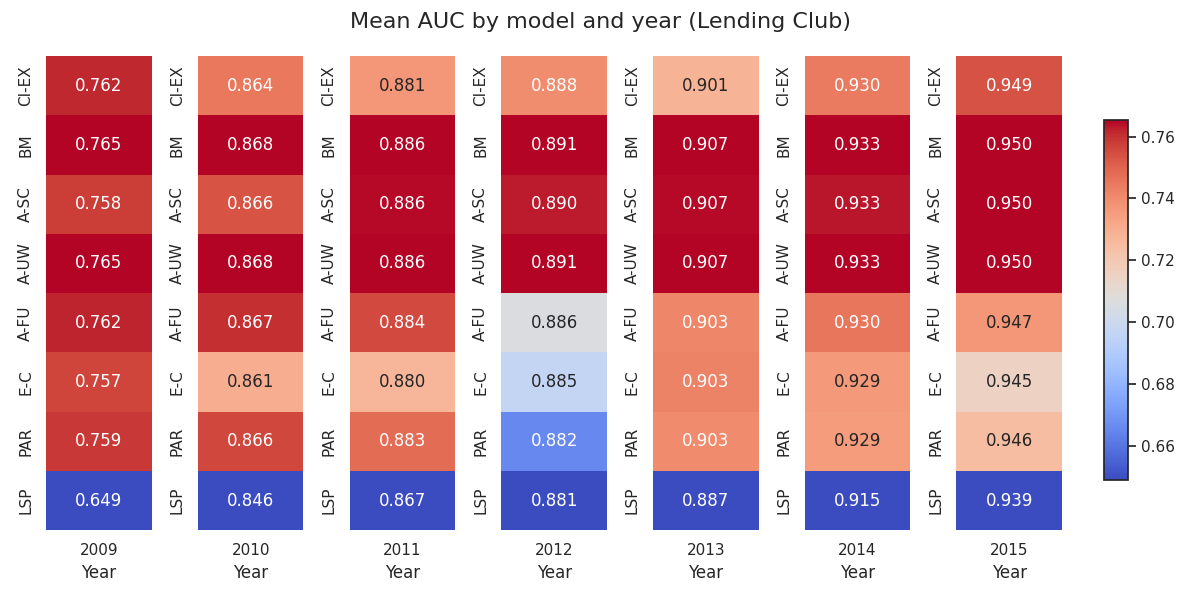}
        \caption{Heatmap illustrating the AUC performance of various models over multiple years for the Lending Club Dataset.}
        \label{fig:auc-heatmap-lc}
    \end{figure}
\begin{figure}
        \centering
    \includegraphics[width=1\linewidth]{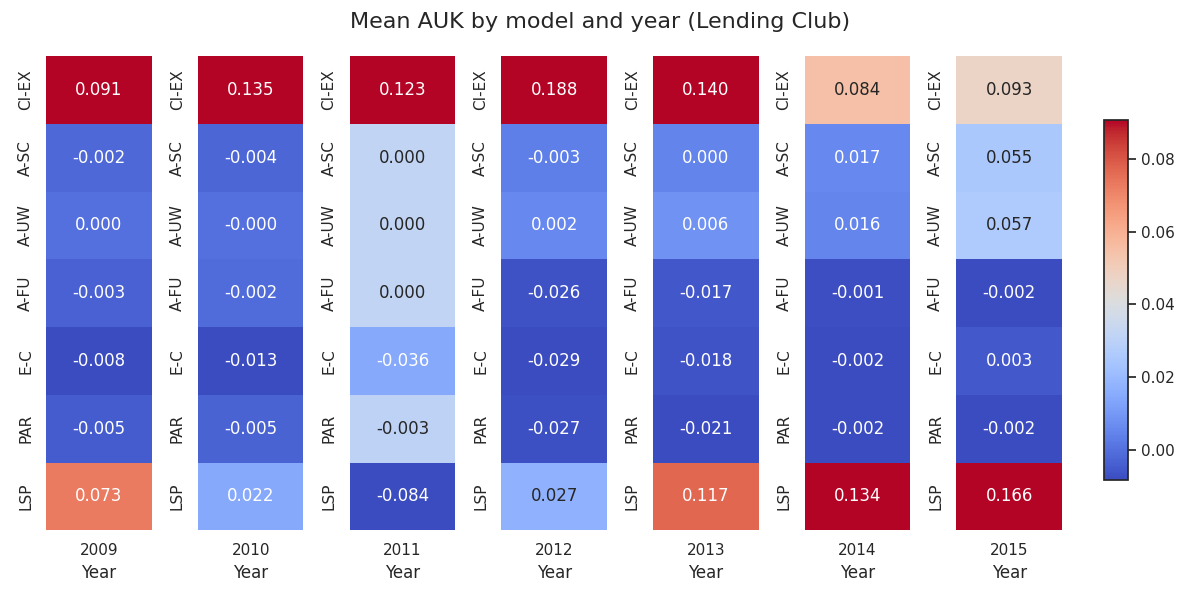}
    \caption{Heatmap displaying the AUK performance across different models and years for the Lending Club Dataset.}
    \label{fig:auk-heatmap-lc}
\end{figure}
\begin{figure}
    \centering
    \includegraphics[width=0.9\linewidth]{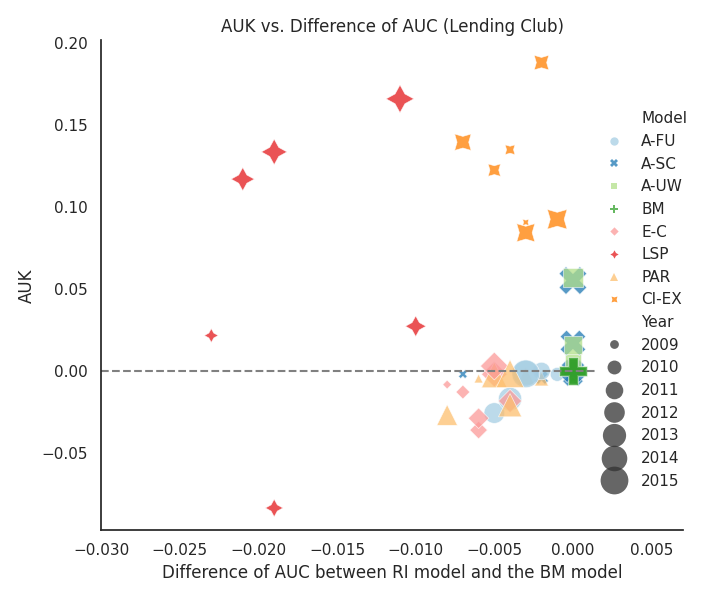}
    \caption{Multi-objective plot showing the relationship between AUC and AUK values for each model for the Lending Club Dataset.}
    \label{fig:auc-auk-lc}
\end{figure}

{From \Cref{fig:auc-heatmap-lc}, we observe that, for most years, the benchmark model (BM) offers the highest AUC performance compared to the RI methods. However, RI techniques such as A-SC, A-UW, and our proposed CI-EX method have achieved comparable AUC scores for many years, with AUC values within less than $1\%$ of the BM model. This similarity is likely due to the reliance of several RI methods on the BM model for labeling rejected samples, creating a strong correlation in AUC values between the two. In contrast, the LSP method, which employs an independent labeling strategy, shows a consistent improvement trend over the years, though its AUC values remain the lowest.}

{The \Cref{fig:auk-heatmap-lc} presents results that contrast with those shown in \Cref{fig:auc-heatmap-lc}. Notably, the LSP method, which consistently showed the lowest AUC values across all years, achieved the highest AUK results in the latter years of the experiment. However, LSP was not the only model with contrasting performance. As the figure indicates, only our proposed CI-EX method performed well across both metrics, achieving the highest AUK scores for most years.}

\subsection{Results Using Simulated Rejected Clients}

We tried to mitigate the lack of available public datasets with information on rejected clients by simulating an accept/reject policy on HomeCredit, which only includes accepted credit applications. This experiment explored how varying levels of strictness in approval and decline policies impact the performance of reject inference (RI) models. The primary distinction between each policy lay in the threshold used to determine group membership: samples with a probability of default above the threshold were classified as "rejects," with their labels disregarded. In contrast, the remaining samples were grouped as "accepted."

For this experiment, as in the previous one with the Lending Club dataset, we presented our results using the same type of plots. Accordingly, we present in \Cref{fig:auc-heatmap-hc} and \Cref{fig:auk-heatmap-hc} the heatmaps of AUC and AUK for each model across different threshold policies, respectively. Also, \Cref{fig:auc-auk-hc} presents a multi-objective perspective for the RI models and both metrics. 

\begin{figure}
        \centering
        \includegraphics[width=1\linewidth]{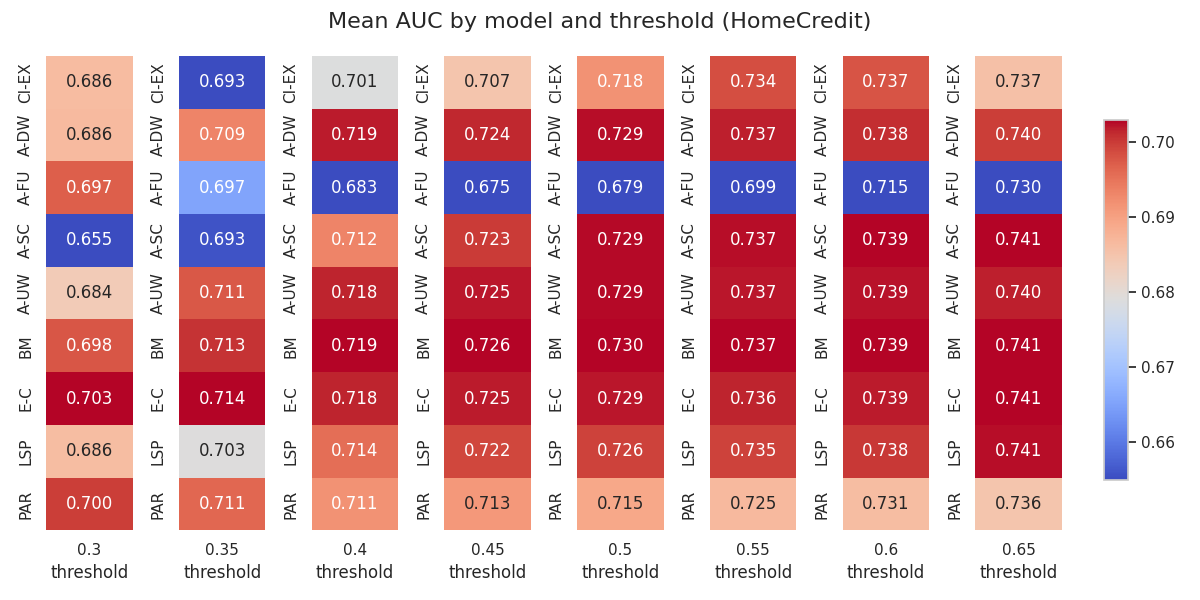}
        \caption{Heatmap illustrating the AUC performance of various models over different thresholds for the HomeCredit Dataset.}
        \label{fig:auc-heatmap-hc}
    \end{figure}
\begin{figure}
        \centering
    \includegraphics[width=1\linewidth]{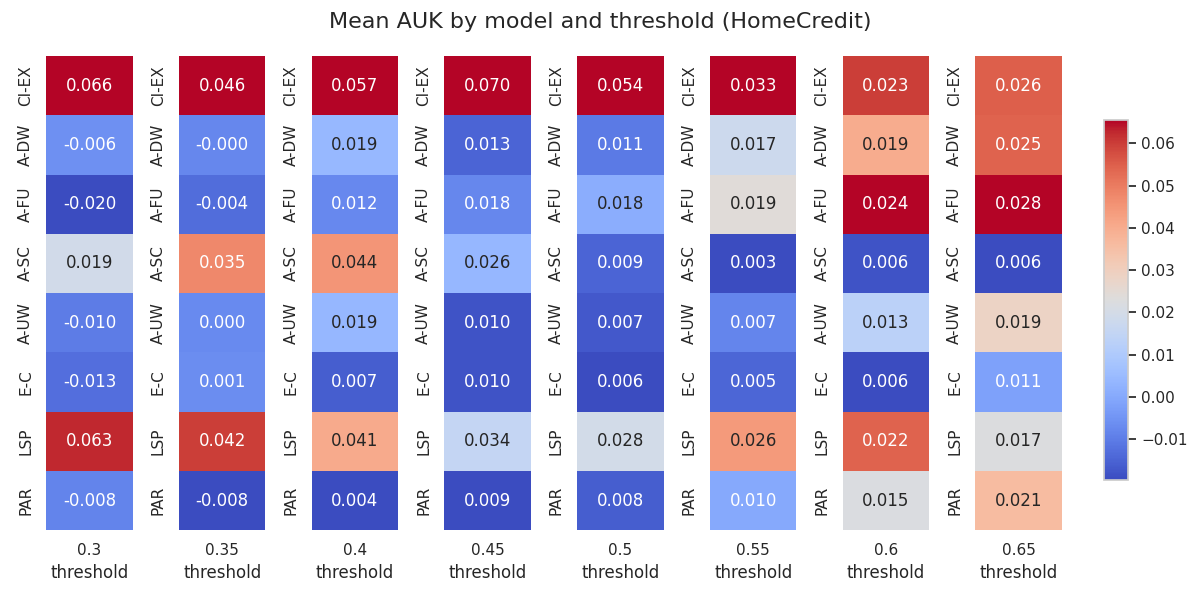}
    \caption{Heatmap illustrating the AUK performance of various models over different thresholds for the HomeCredit Dataset.}
    \label{fig:auk-heatmap-hc}
\end{figure}
\begin{figure}
    \centering
    \includegraphics[width=0.9\linewidth]{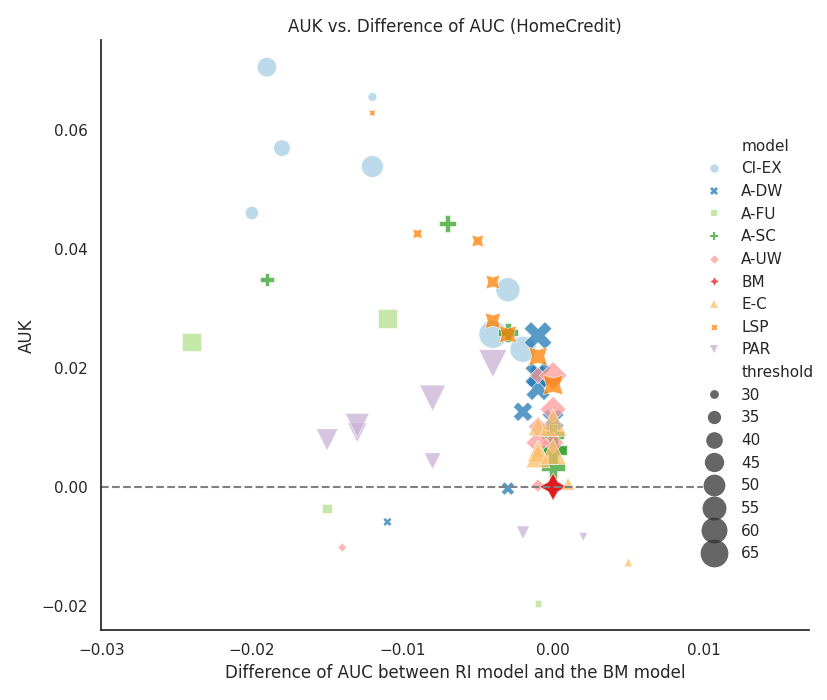}
    \caption{Multi-objective plot showing the relationship between AUC and AUK values for each model for the HomeCredit Dataset.}
    \label{fig:auc-auk-hc}
\end{figure}

{As in the previous experiment, \Cref{fig:auc-heatmap-hc} provides a comparative overview of how various RI models perform relative to the Benchmark model. Notably, the extrapolation method (E-C), which closely resembles our proposed CI-EX, outperforms the Benchmark model under certain threshold policies. Interestingly, unlike the experiment using the LC dataset, fuzzy augmentation consistently ranks as the lowest-performing method according to the AUC metric in this dataset.}

{For threshold values greater than $0.4$, the performance ranking among methods remains relatively stable. However, we observe a notable correlation between threshold values and AUC scores, particularly for our CI-EX method, which becomes more pronounced as thresholds increase. This trend may be attributable to the models’ access to a larger number of labeled training samples, as illustrated in \Cref{fig:accept-reject-split}.}

{In \Cref{fig:auk-heatmap-hc}, we can see the performance of the RI models on the AUK metric for each threshold. The figure indicates that our proposed CI-EX method outperforms most RI methods on this RI-specific metric. For CI-EX, there is an inverse correlation between threshold values and AUK scores, with higher thresholds resulting in lower AUK values. In contrast, other techniques, such as E-C and A-FU, display a direct correlation, showing higher AUK scores with increasing threshold values.}

\begin{figure}
    \centering
    \includegraphics[width=1\linewidth]{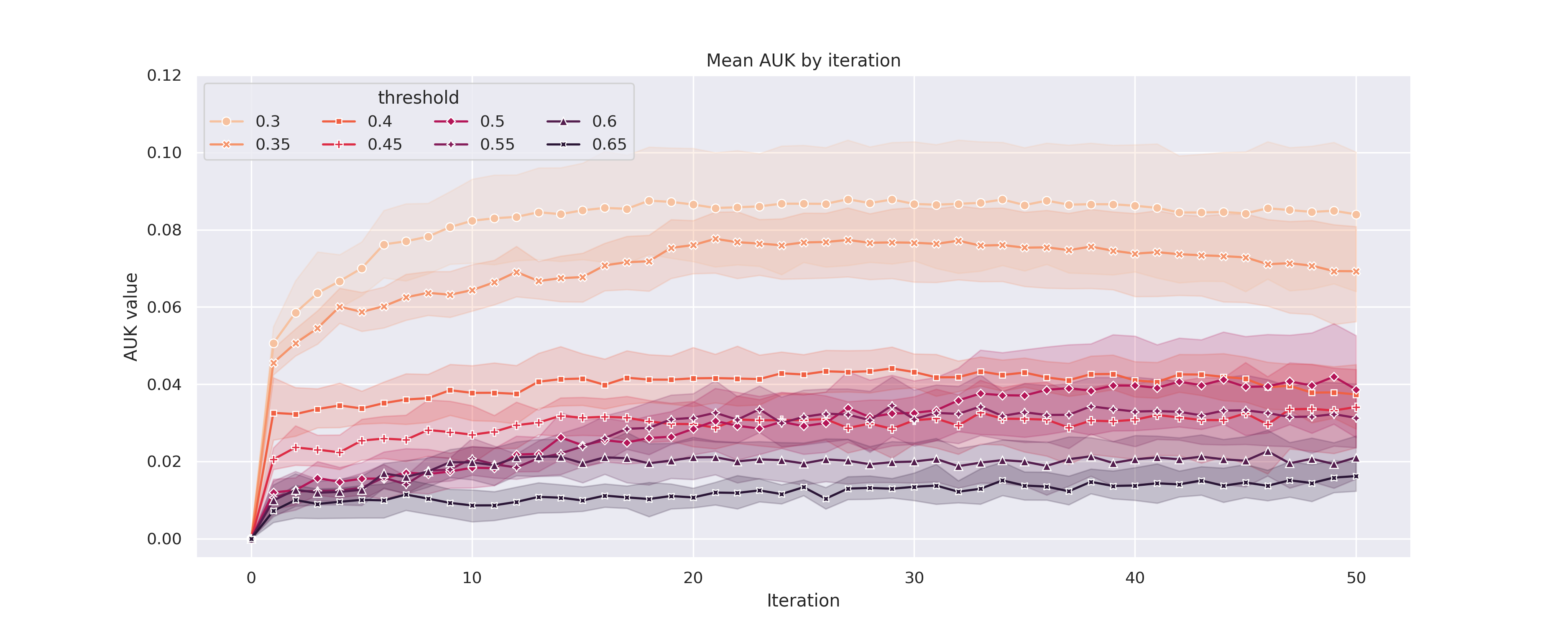}
    \caption{AUK evolution by iteration of CI-EX technique for different thrsehold policies in HomeCredit dataset}
    \label{fig:hc-iterations-auk}
\end{figure}

\begin{figure}
    \centering
    \includegraphics[width=1\linewidth]{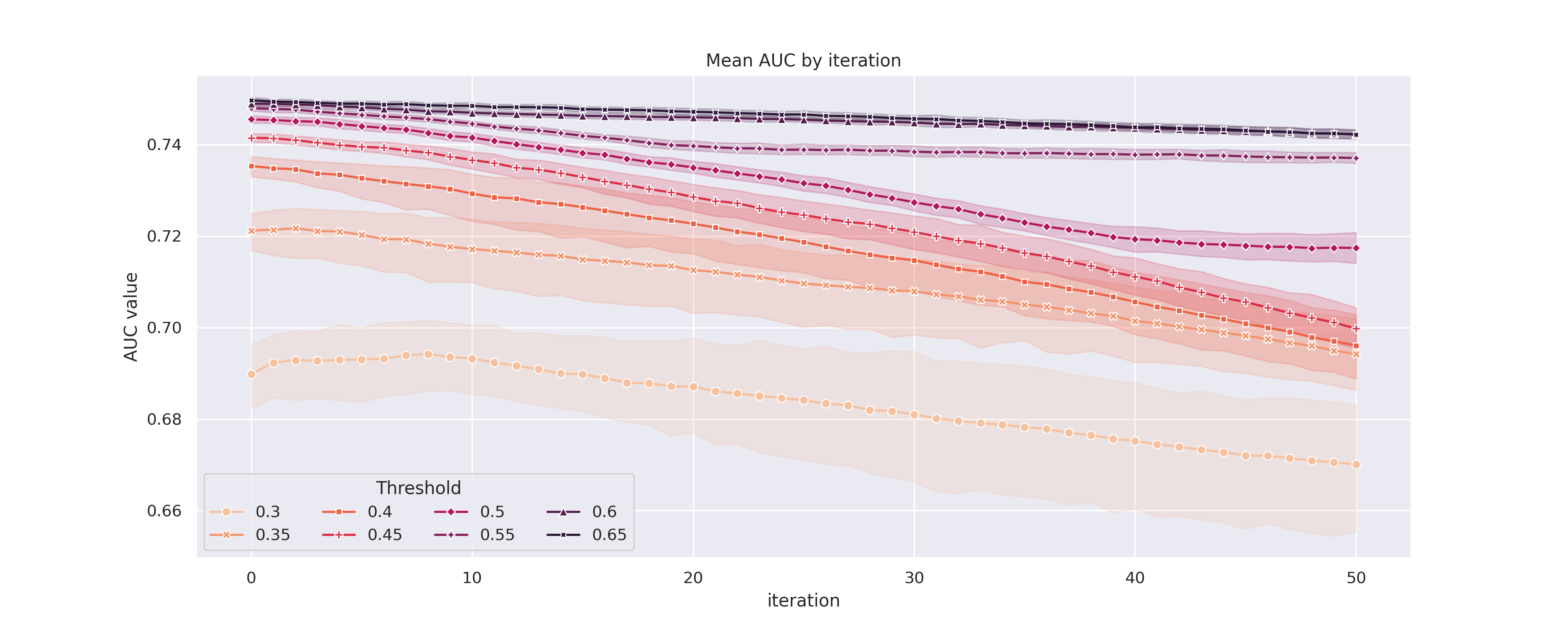}
    \caption{AUC evolution by iteration of CI-EX technique for different threshold policies in HomeCredit dataset}
    \label{fig:hc-iterations-auc}
\end{figure}

\section{Discussion}

As shown in \Cref{fig:auc-auk-lc}, which combines both metrics in a multi-objective plot. There is clearly some trade-offs between AUC and AUK across models. The Y-axis represents the AUK values, while the X-axis shows the difference in AUC between each RI model and the BM model for each year, and each point represents a different model. Most of the models fall in the region where AUK values are near or mostly below zero, indicating that these strategies offer limited improvements over the BM model. Two notable outliers emerge: our proposed CI-EX model, which occupies the region with positive results in both metrics, and the LSP model, which achieves high AUK values in certain cases but consistently low AUC scores.

\Cref{fig:auc-auk-hc} presents both metrics in a multi-objective plot. As in the previous experiment with the Lending Club dataset, the Y-axis represents the AUK values, while the X-axis shows the AUC difference between each RI model and the Benchmark model at each threshold. Each point in the plot corresponds to a different model. Unlike the previous experiment, most models are located in a region with positive AUK values. Figures \Cref{fig:auk-heatmap-hc} and \Cref{fig:auc-auk-hc} also show that models with negative AUK values are associated with runs using smaller threshold values. However, our proposed model does not follow this trend and maintains stability across different thresholds. The plot reveals that few models achieve high values for both metrics simultaneously across various policies in this dataset, indicating a potential trade-off in optimizing AUK and AUC metrics.

One of our concerns was establishing the chosen iteration of our proposed model for comparison with other RI models. As mentioned in the methodology, we used the TOPSIS tool to make this selection. However, alternative selection methods could also be applied, such as input from credit specialists or product owners. In \Cref{fig:hc-iterations-auc} and \Cref{fig:hc-iterations-auk}, we present the mean AUC and AUK values, along with confidence intervals, for each iteration of our model across different thresholds. These figures show that specific iterations of the CI-EX method can achieve higher AUK values while experiencing smaller losses in AUC compared to the results shown in previous plots. The AUK and AUC metrics exhibit a strong correlation with iteration number, with AUK showing a positive correlation and AUC showing a negative one. From the figures, the "sweet spot" for this method appears to lie around iterations 10 to 20, where the gains in AUK outweigh the relative losses in AUC. However, this spot could vary depending on the size and structure of the datasets.

\section{Conclusions and future work}
This research successfully addressed the challenges posed by reject inference in credit scoring, confirming the relevance of the hypotheses outlined in the paper as hypothesized: (1) Relying solely on data from accepted clients despite having high AUC proved insufficient to deal with reject data (low AUK), underscoring the need to incorporate information about rejected applicants. (2) Significant differences exist between the distributions of accepted and rejected clients. CI-EX effectively addressed this challenge by leveraging outlier detection and a confidence-based selection criterion to incorporate information from rejected clients iteratively. (3) It is possible to infer the behavior of rejected clients by utilizing distributional information and appropriate techniques, as CI-EX successfully captured valuable insights about rejected applicants. At the same time, other methods do not handle this distributional information interactively and usually struggle to find good models. (4) Finally, evaluating RI models exclusively on accepted client data fails to accurately reflect their true performance, highlighting the necessity of employing appropriate metrics and evaluation strategies like the proposed AUK metric.

This research culminated in CI-EX, a valuable solution for reject inference in credit scoring that directly addresses the key hypotheses outlined. By effectively leveraging accepted client data and carefully extrapolating to the rejected population, CI-EX contributes to a fairer and more inclusive credit assessment.  Our evaluation demonstrated that CI-EX uniquely excels at both the established AUC metric and the novel RI-specific AUK metric, highlighting the importance of AUK as a complementary measure. CI-EX achieves this by strategically optimizing AUK, accepting marginal trade-offs in AUC to achieve superior performance compared to other RI techniques. This Pareto-optimal approach positions CI-EX as the leading method for improving AUK, which is particularly crucial given the inherent conflict between maximizing AUC and AUK in reject inference.  Furthermore, utilizing the kick-out metric, our findings challenge previous conclusions by demonstrating that even classical RI techniques can enhance model performance.

Despite these contributions, our work has limitations: CI-EX requires longer training times than other RI methods, and experiments were conducted using a single type of classifier model. Future research should address these limitations by optimizing CI-EX for efficiency, exploring alternative classifier models, applying the framework to various credit datasets, and investigating metrics that assess the impact of RI methods on marginalized populations, as well as refining the AUK metric.

\section*{Acknowledgments:}
This project was supported by the Brazilian Ministry of Science, Technology and Innovations, with resources from Law nº 8,248, of October 23, 1991, within the scope of PPI-SOFTEX, coordinated by Softex and published Arquitetura Cognitiva (Phase 3), DOU 01245.003479/2024 -10.

\section*{CRediT authorship contribution statement}

\textbf{Athyrson M. Ribeiro:} Conceptualization, Methodology, Software, Writing. \textbf{Marcos Medeiros Raimundo:} Review and Editing.

\section*{Declaration of Interests:} 

The authors declare that they have no known competing financial interests or personal relationships that could have appeared to influence the work reported in this paper.

\section*{Data Availability Statement}

The datasets analyzed in the current study are publicly available in the Kaggle repositories: Home Credit Default Risk dataset \cite{homecredit} and Lending Club Loan Data dataset \cite{lendingclub}. No additional datasets were generated or analyzed during the current study.

\newpage



\bibliographystyle{elsarticle-num}

\bibliography{ref}

\newpage

\appendix

\section{Hyper-Parameter optimization}
\label{app1}

\begin{figure}[htb]
  \centering
    \includegraphics[width=\linewidth]{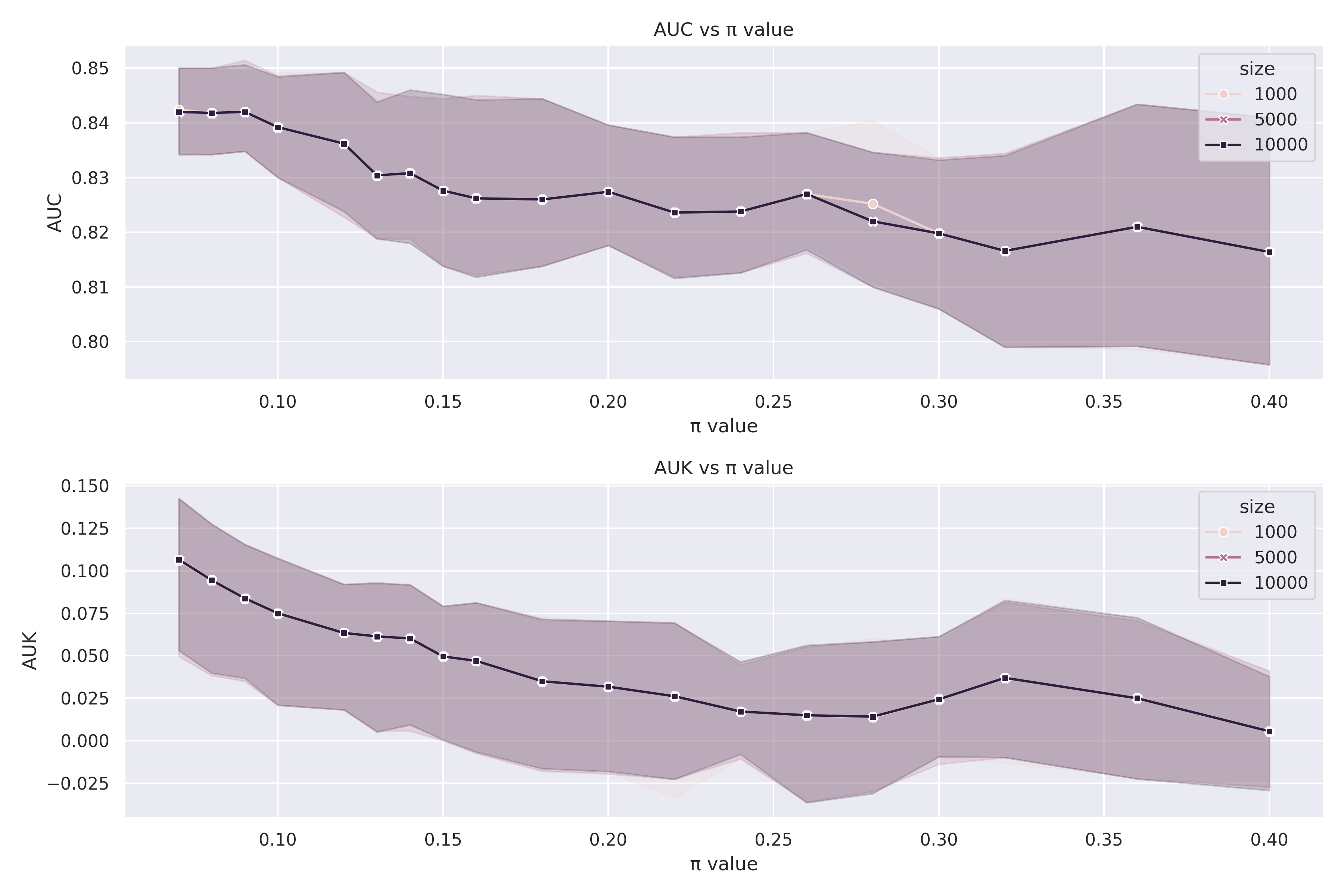}
    \caption{Metrics relation with parameters $\pi$ and $\eta$}
  \label{fig:auk_pi_etha}

\end{figure}


\section{Tables}

\begin{sidewaystable}[h!]
\centering
\footnotesize
  \caption{Description of $S_1$ group of Homecredit features}
  \label{table:hcdesc}
  \begin{tabular}{cl p{6cm}}
    \toprule
    ID & Features & Description\\
    \midrule
    F1 & AMT\_CREDIT & Credit amount of the loan \\
    F2,F3 and F4 & EXT\_SOURCE\_1,2 and 3 & Normalized score from external data sources \\
    F5 & REGION\_POPULATION\_RELATIVE & Normalized population of region where client lives \\
    F6 & DAYS\_EMPLOYED & How many days before the application the person started current employment \\
    F7 & DAYS\_BIRTH & Client's age in days at the time of application \\
    F8 & AMT\_INCOME\_TOTAL & Income of the client \\
    F9 & CNT\_CHILDREN & Number of children the client has \\
    F10 & CNT\_FAM\_MEMBERS & How many family members does the client have \\
    F11 & REG\_CITY\_NOT\_WORK\_CITY & Flag if client's permanent address does not match work address \\
    F12 & AMT\_GOODS\_PRICE & For consumer loans it is the price of the goods for which the loan is given \\
    F13 & ALAG\_OWN\_CAR & Flag if the client owns a car \\
    F14 & NAME\_EDUCATION\_TYPE & Level of highest education the client achieved \\
    F15 & NAME\_CONTRACT\_TYPE & Identification if loan is cash or revolving \\
    F16 & TARGET & Target variable (1 - client with payment difficulties: he/she had late payment more than X days on at least one of the first Y installments of the loan in our sample, 0 - all other cases)\\
    \bottomrule
  \end{tabular}
\end{sidewaystable}

\begin{table*}[hbp]
  \centering
  \footnotesize
  \caption{Description of features for accepted clients dataset}
  \label{table:lcdesc-acp}
  \begin{tabular}{l l p{8cm}}
    \toprule
    ID & Features & Description \\
    \midrule
    A1 & addr\_state & The state the borrower provides in the loan application. \\
    A2 & annual\_inc & The self-reported annual income provided by the borrower during registration. \\
    A3 & delinq\_2yrs & The number of 30+ days past-due incidences of delinquency in the borrower's credit file for the past 2 years. \\
    A4 & dti & A ratio is calculated using the borrower’s total monthly debt payments on the total debt obligations, excluding mortgage and the requested LC loan, divided by the borrower’s self-reported monthly income. \\
    A5 & emp\_length & Employment length in years. \\
    A6 & home\_ownership & The homeownership status provided by the borrower during registration or obtained from the credit report. \\
    A7 & int\_rate & Interest Rate on the loan. \\
    A8 & issue\_d & The month in which the loan was funded. \\
    A9 & last\_fico\_range\_high & The upper boundary range the borrower’s last FICO pulled belongs to. \\
    A10 & last\_fico\_range\_low & The lower boundary range the borrower’s last FICO pulled belongs to. \\
    A11 & inq\_last\_6mths & The number of inquiries in the past 6 months (excluding auto and mortgage inquiries). \\
    A12 & loan\_amnt & The listed loan amount applied for by the borrower.\\ 
    A13 & revol\_util & Revolving line utilization rate, or the amount of credit the borrower uses relative to all available revolving credit. \\
    A14 & term & The number of payments on the loan. Values are in months and can be either 36 or 60. \\
    \hline
    A15 & loan\_status & Current status of the loan. \\ 
    \bottomrule
  \end{tabular}
\end{table*}

\begin{table*}[hbp]
  \centering
  \footnotesize
  \caption{Description of features for rejected clients dataset}
  \label{table:lcdesc-rej}
  \begin{tabular}{l l p{8cm}}
    \toprule
    ID & Features & Description \\
    \midrule
    R1 & Amount Requested & The total amount requested by the borrower. \\
    \addlinespace
    R2 & Application Date & The date on which the borrower applied. \\
    \addlinespace
    R3 & Risk\_Score & For applications before November 5, 2013, the risk score is the borrower's FICO score. For applications after November 5, 2013, the risk score is the borrower's Vantage score. \\
    \addlinespace
    R4 & Debt-To-Income Ratio & A ratio calculated using the borrower’s total monthly debt payments on the total debt obligations, excluding mortgage and the requested LC loan, divided by the borrower’s self-reported monthly income. \\
    \addlinespace
    R5 & State & The state provided by the borrower in the loan application. \\
    \addlinespace
    R6 & Employment Length & Employment length in years. \\
    \bottomrule
  \end{tabular}
\end{table*}


\begin{table*}[hbp]
\centering
\caption{Number of Accepted and Rejected Samples for Different Thresholds ($\epsilon$)}
\begin{tabular}{c|ccc|ccc}
\hline
$\epsilon$ & \multicolumn{3}{c|}{Accepted Samples} & \multicolumn{3}{c}{Rejected Samples} \\ 
\hline
           & Train  & Test   & Validation & Train   & Test   & Validation \\ 

0.30       & 7463   & 2980   & 2343       & 118492  & 46222  & 37019      \\ 
0.35       & 17595  & 6850   & 5462       & 108360  & 42352  & 33900      \\ 
0.40       & 32408  & 12789  & 10089      & 93547   & 36413  & 29273      \\ 
0.45       & 52600  & 20616  & 16358      & 73355   & 28586  & 23004      \\ 
0.50       & 72808  & 28560  & 22646      & 53147   & 20642  & 16716      \\ 
0.55       & 93630  & 36677  & 29166      & 32325   & 12525  & 10196      \\ 
0.60       & 108072 & 42250  & 33657      & 17883   & 6952   & 5705       \\ 
0.65       & 117612 & 45884  & 36699      & 8343    & 3318   & 2663       \\ 
\hline
\end{tabular}
\label{tab:number_accept_reject_split}
\caption*{The number of samples classified into the Accepted and Rejected groups for each subset, given the threshold value.}
\end{table*}





\end{document}